\documentclass{article} 
\usepackage{iclr2027_conference,times}

\usepackage{amsmath,amsfonts,bm}

\def\eqref#1{equation~\ref{#1}}

\def\1{\bm{1}}

\DeclareMathAlphabet{\mathsfit}{\encodingdefault}{\sfdefault}{m}{sl}
\SetMathAlphabet{\mathsfit}{bold}{\encodingdefault}{\sfdefault}{bx}{n}

\usepackage{hyperref}
\usepackage{url}
\usepackage{tikz}
\usetikzlibrary{arrows.meta,positioning,calc,backgrounds,shapes.geometric,decorations.pathreplacing,fit}
\usepackage{algorithm}
\usepackage{algorithmic}
\usepackage{amssymb}
\usepackage{graphicx}
\usepackage{booktabs}
\usepackage{multirow}
\usepackage{wrapfig}
\usepackage{placeins}
\usepackage{flafter}

\title{Dual-Context Analog Retrieval for\\
Time-Series Forecasting}

\author{Jung Min Choi, Ngoc Son Le, Ibram Abdelmalak, \\
\textbf{Vijaya Krishna Yalavarthi \& Lars Schmidt-Thieme} \\
ISMLL, University of Hildesheim \\
VWFS Data Analytics Research Center (VWFS-DARC), Hildesheim, Germany \\
\texttt{choi@ismll.de}, \texttt{sle@ismll.de}, \texttt{abdelmalak@ismll.de}, \\
\texttt{yalavarthi@ismll.de}, \texttt{schmidt-thieme@ismll.de}
}

\iclrfinalcopy
\begin{document}

\maketitle

\begin{abstract}
Most long-term time-series forecasting models map the look-back window directly to the full horizon in a single pass. While efficient, this design does not explicitly determine which historical state is most relevant to each future segment or consider what followed that state. Analog forecasting can provide such evidence by retrieving a past state similar to the present and examining its observed continuation. However, relying on a single nearest match can be unreliable, and overlapping patches may yield highly similar or near-duplicate candidates. To address these limitations, we propose DuoTS, a Dual-Context Time Series forecasting model, which leverages retrieved evidence without depending on it exclusively. DuoTS first generates a base forecast using a fully parallel patch encoder and a linear prediction head, and then progressively refines this forecast one future patch at a time. Each refinement considers two complementary views. The first view, the current context, attends to recent tokens and captures the latest dynamics of the series. The second view, the detail context, provides distinct retrieved analogs together with the trajectories that followed them. Since the refinement is performed separately for each future patch, the model weighs the two views differently across the forecast horizon and associates each segment of the prediction with the evidence most appropriate to its temporal distance from the present. Extensive experiments on multiple real-world time-series datasets show that DuoTS achieves state-of-the-art performance, while component-wise ablation studies verify the contribution of each context. Moreover, the refinement mechanism is independent of the remaining DuoTS architecture. It requires only the encoded look-back window and the position of the future patch, allowing it to be applied as a model-agnostic module to improve existing forecasting models.

\end{abstract}

\section{Introduction}
\label{sec:introduction}

\begin{wrapfigure}{r}{0.47\textwidth}
\centering
\vspace{-13pt}

\definecolor{lcsignal}{RGB}{104,120,181}
\definecolor{lcgreenfill}{RGB}{223,240,224}
\definecolor{lcgreenborder}{RGB}{86,140,90}
\definecolor{lcorangefill}{RGB}{252,230,204}
\definecolor{lcorangeborder}{RGB}{198,124,44}
\definecolor{lccreamfill}{RGB}{250,238,213}
\definecolor{lccreamborder}{RGB}{193,148,72}
\definecolor{lcrevisedborder}{RGB}{118,86,160}

\resizebox{\linewidth}{!}{%
\begin{tikzpicture}[
  x=1cm,y=1cm,
  font=\sffamily\small,
  >=Latex,
  every node/.style={inner sep=1pt,outer sep=1pt},
  panel/.style={draw=black!22,fill=black!1,rounded corners=5pt,line width=0.7pt},
  panelshadow/.style={draw=none,fill=black!13,rounded corners=5pt},
  title/.style={font=\sffamily\normalsize\bfseries,text=black!78},
  note/.style={font=\sffamily\footnotesize,text=black!58,align=center},
  view/.style={rounded corners=2pt,line width=0.7pt,minimum height=5.5mm,
               inner xsep=3pt,font=\sffamily\scriptsize\bfseries},
  evidenceplus/.style={font=\sffamily\footnotesize\bfseries,text=black!45},
  flow/.style={-Latex,line width=0.9pt,draw=black!55},
]

\node[panelshadow,minimum width=9.15cm,minimum height=1.95cm] at (4.79,5.85) {};
\node[panel,minimum width=9.15cm,minimum height=1.95cm] at (4.75,5.90) {};
\node[title,anchor=west] at (0.35,6.62) {(a) One-shot forecast};

\draw[black!32,line width=0.55pt] (0.55,5.24) -- (9.02,5.24);
\draw[black!36,dashed,line width=0.65pt] (6.18,5.04) -- (6.18,6.35);
\node[note,anchor=north,font=\sffamily\scriptsize] at (2.95,5.15) {observed history};
\node[note,anchor=north,font=\sffamily\scriptsize] at (7.70,5.15) {future};

\draw[lcsignal,line width=1.35pt]
  plot[smooth] coordinates {
    (0.62,5.56) (1.02,5.98) (1.42,5.54) (1.82,6.08)
    (2.22,5.63) (2.62,6.19) (3.02,5.72) (3.42,5.43)
    (3.82,5.91) (4.22,5.50) (4.62,6.04) (5.02,5.59)
    (5.42,6.12) (5.82,5.67) (6.18,5.46)
  };
\draw[black!48,line width=1.15pt,dash pattern=on 3pt off 2pt]
  plot[smooth] coordinates {
    (6.18,5.46) (6.72,5.61) (7.26,5.74) (7.80,5.81)
    (8.34,5.86) (8.90,5.90)
  };
\node[note,text=black!55,anchor=south] at (7.78,5.93) {predicted forecast};

\node[panelshadow,minimum width=9.15cm,minimum height=3.40cm] at (4.79,3.05) {};
\node[panel,minimum width=9.15cm,minimum height=3.40cm] at (4.75,3.10) {};
\node[title,anchor=west] at (0.35,4.55)
  {(b) DuoTS: \textcolor{lcgreenborder}{current} $+$
   \textcolor{lcorangeborder}{detail} view revision};

\draw[black!32,line width=0.55pt] (0.55,2.85) -- (9.02,2.85);
\draw[black!36,dashed,line width=0.65pt] (6.18,2.62) -- (6.18,4.30);

\foreach \x in {6.89,7.60,8.31}{
  \draw[black!14,line width=0.55pt,dash pattern=on 1pt off 1.2pt]
    (\x,1.98) -- (\x,3.95);
}

\draw[lcsignal,line width=1.20pt]
  plot[smooth] coordinates {
    (0.62,3.17) (1.02,3.59) (1.42,3.15) (1.82,3.69)
    (2.22,3.24) (2.62,3.80) (3.02,3.33) (3.42,3.04)
    (3.82,3.52) (4.22,3.11) (4.62,3.65) (5.02,3.20)
    (5.42,3.73) (5.82,3.28) (6.18,3.07)
  };

\draw[black!42,line width=1.05pt,dash pattern=on 3pt off 2pt]
  plot[smooth] coordinates {
    (6.18,3.07) (6.72,3.22) (7.26,3.35) (7.80,3.42)
    (8.34,3.47) (8.90,3.51)
  };

\draw[lcorangeborder,line width=2.15pt]
  plot[smooth] coordinates {(0.70,3.23) (1.02,3.59) (1.42,3.15)};
\draw[lcorangeborder,line width=1.25pt,dash pattern=on 2pt off 1.4pt,-Latex]
  plot[smooth] coordinates {(1.42,3.15) (1.82,3.69) (2.22,3.24)};

\draw[lcorangeborder,line width=2.15pt]
  plot[smooth] coordinates {(3.50,3.12) (3.82,3.52) (4.22,3.11)};
\draw[lcorangeborder,line width=1.25pt,dash pattern=on 2pt off 1.4pt,-Latex]
  plot[smooth] coordinates {(4.22,3.11) (4.62,3.65) (5.02,3.20)};

\draw[lcgreenborder,line width=2.15pt,-{Latex[length=2.2mm,width=1.8mm]}]
  plot[smooth] coordinates {(5.18,3.41) (5.42,3.73) (5.82,3.28) (6.18,3.07)};

\draw[lcrevisedborder,line width=1.75pt]
  plot[smooth] coordinates {
    (6.18,3.07) (6.72,3.64) (7.26,3.21) (7.80,3.75)
    (8.34,3.32) (8.90,3.62)
  };
\node[note,text=lcrevisedborder,anchor=south,font=\sffamily\footnotesize\bfseries]
  at (7.95,3.99) {revised};

\foreach \xc/\hg in {6.535/0.42,7.245/0.32,7.955/0.22,8.665/0.14}{
  \draw[lcgreenborder,fill=lcgreenborder!55,line width=0.5pt]
    (\xc-0.19,2.00) rectangle (\xc+0.19,2.00+\hg);
  \draw[lcorangeborder,fill=lcorangeborder!55,line width=0.5pt]
    (\xc-0.19,2.00+\hg) rectangle (\xc+0.19,2.58);
}
\node[note,anchor=north] at (7.60,1.92) {view weights per patch};

\node[view,draw=lcgreenborder,fill=lcgreenfill,text=lcgreenborder,
      anchor=west] (current) at (1.05,2.30) {Current view};
\node[evidenceplus,right=1.5mm of current] (plusdetail) {$+$};
\node[view,draw=lcorangeborder,fill=lcorangefill,text=lcorangeborder,
      right=1.5mm of plusdetail] (detail) {Detail view};

\end{tikzpicture}%
}
\vspace{-6pt}
\caption{\small \textbf{Forecast, retrieve, revise.} (a) One-shot forecasting
predicts the full horizon directly. (b) DuoTS revises each future patch using
current and retrieved-detail evidence with learned view weights.}
\label{fig:lc6_thumbnail}
\vspace{-12pt}
\end{wrapfigure}
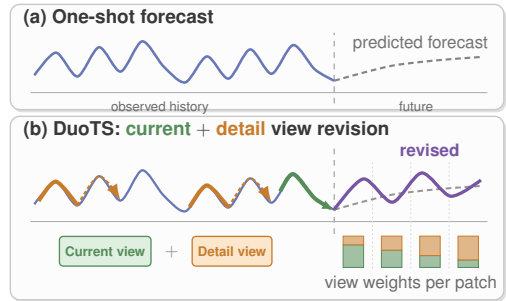

Forecasting models predict the future from the past, and much of what they
exploit is recurring structure: trends that persist and cycles that repeat.
Models differ in how they use this structure. One class learns it implicitly.
PatchTST, iTransformer, and TSMixer map the look-back window directly to the
horizon and achieve strong accuracy without an explicit notion of trend or
cycle \citep{patchtst,itransformer,tsmixer}. A second class separates recurring
structure explicitly. Decomposition models such as Autoformer, DLinear, and
TimeMixer split or mix trend and seasonal components
\citep{autoformer,dlinear,timemixer}, while CycleNet learns a recurrent cycle
for each channel and adds it back to the prediction \citep{cyclenet}.

Both classes rely on structure learned across the training set. An implicit
model must discover for itself which past states matter for a given window,
whereas an explicit cycle describes a typical period and stays fixed after
training. Neither is designed to use what the current window itself reveals.
Consider electricity demand during a heat wave. A learned daily cycle describes
a typical day, but a heat-wave day is not typical. The most relevant evidence
is the earlier days of the same heat wave and how demand evolved after them,
and the window already holds it. This is the principle of analog forecasting:
find a past state resembling the present and use what followed it
\citep{lorenz1969,vandendool1994}.

We propose \textbf{DuoTS}, which consists of a base forecaster and a corrector.
The base forecaster is a fully patch-based model that mixes information
sparsely across channels and across time through local dynamics. The corrector
makes window-specific evidence explicit and uses it to refine each future patch
of the base prediction. It looks at the past through two views: the
\emph{current context}, which attends to recent tokens and captures what the
series has just been doing, and the \emph{detail context}, which retrieves earlier states resembling the
present together with what followed them. Retrieved analogs can nevertheless
be misleading: the nearest match is not always the most predictive,
overlapping patches yield near-duplicates, and sometimes no good analog exists.
The corrector therefore refines the base prediction rather than overriding it,
adjusting it only as far as the retrieved evidence warrants. Figure~\ref{fig:lc6_thumbnail}
contrasts this forecast--retrieve--revise process with one-shot forecasting.

The corrector reads only patch embeddings, so it can be attached to other
patch-based forecasters without changing its design. Attached to SRSNet,
TimeKAN, and PatchTST, it improves 15 of 24 tested settings under joint
training. The full DuoTS model achieves the lowest MSE in 22 and the lowest MAE
in 23 of 32 dataset--horizon settings across eight benchmarks.

Our contributions are as follows.
\begin{itemize}
    \item We propose DuoTS, which pairs a base forecaster with a corrector that
    makes window-specific evidence explicit through two views of the past:
    recent dynamics and retrieved analogs.
    \item DuoTS achieves state-of-the-art results on eight long-term forecasting
    benchmarks, obtaining the lowest MSE in 22 and the lowest MAE in 23 of 32
    dataset--horizon settings.
    \item The corrector is model-agnostic: it requires only patch embeddings and
    transfers across backbones. Attached to SRSNet, TimeKAN, and PatchTST, it
    improves accuracy in 15 of 24 tested settings. Code is available at
    \url{https://anonymous.4open.science/r/DuoTS-244D}.
\end{itemize}

\section{Related Work}
\label{sec:related}

\paragraph{Time-series forecasting.}
Modern forecasting models use several architectural families. DLinear maps
decomposed signal components through linear layers \citep{dlinear}. TSMixer and
TimeMixer use MLP mixing over temporal features or temporal scales
\citep{tsmixer,timemixer}. PatchTST represents temporal segments as tokens while
iTransformer instead represents variables as tokens
\citep{patchtst,itransformer}. Recent models further explore wavelet features
in SimpleTM \citep{chen2025simpletm} and frequency decomposition in TimeKAN
\citep{timekan}. SRSNet learns which patch representations to retain
\citep{srsnet}. The lightweight TSMixer of \citet{tsmixer2} is the closest
backbone-level precedent to DuoTS: it applies MLP mixing to patched multivariate
series and explicitly models channel interactions with gated components. DuoTS
instead combines a direct patch path with a structured path containing sparse
cross-channel mixing and explicit first- and second-order patch dynamics, then
adds a separate evidence query and dual-context correction for each future
patch.

\paragraph{Retrieval-augmented forecasting.}
Classical analog forecasting uses the continuation of a similar past state
\citep{lorenz1969,vandendool1994}. Context Parroting demonstrates that related
copying strategies remain competitive for scientific time series
\citep{contextparrot}. Recent methods retrieve evidence from training data or a
global memory. RATD guides a diffusion model with retrieved references
\citep{ratd}. RAFT supplies matched historical futures \citep{raft}. TS-RAG
augments a pretrained foundation model with retrieved patterns
\citep{tsrag}. Other methods use a cached nearest-neighbor store
\citep{knnmts} or retrieve aligned global segments \citep{gtr}. PFRP combines
global retrieval with a local model \citep{pfrp}. SARAF promotes both relevance
and diversity under nonstationarity \citep{saraf}. DuoTS instead retrieves from
the current look-back window for every future patch. Each retrieved analog is
paired with its successors. These successors are the trajectories observed
immediately after the matched historical patch. The model places this evidence
beside the current context and uses it to refine a direct forecast.

\paragraph{Multi-resolution history modeling.}
TimeMixer represents a series at several sampling scales \citep{timemixer}.
Pathformer learns pathways across multiple patch sizes \citep{pathformer}.
HDMixer captures dependencies within patches as well as dependencies across
patches and variables \citep{hdmixer}. GRACE groups a long user history into
journeys and reads several scopes of that history \citep{grace}. DuoTS borrows
only the idea of grouping a long sequence into ordered blocks. Those blocks
serve as a coarse search index rather than an additional context branch. The
search first selects promising blocks and then retrieves individual patches
inside them. This design reduces the search space while preserving detailed
analog continuations for the final refinement.

\section{Method}
\label{sec:method}

For one sample, let $\mathbf X\in\mathbb R^{C\times L}$ be a look-back window
of $L$ observations from each of $C$ channels, and let
$\mathbf Y\in\mathbb R^{C\times H}$ contain the next $H$ target values. We
omit the batch dimension throughout. Reversible instance normalization
(RevIN; \citealp{revin}) normalizes each input channel, and
$\operatorname{RevIN}^{-1}$ maps a normalized forecast back to its original
scale. DuoTS produces a normalized base forecast
$\widehat{\mathbf Y}^{\mathrm{base}}\in\mathbb R^{C\times H}$ and refines it
using the dual-context correction
$\widehat{\mathbf E}\in\mathbb R^{C\times H}$. Let $P$ be the patch length and
$F=\lceil H/P\rceil$ the number of future patches. Each future patch $f$ has a
learned gate $\zeta_f=\tanh(\theta_f)$, repeated over its $P$ output positions
to form $\overline{\boldsymbol\zeta}\in\mathbb R^H$. The final forecast is
\begin{equation}
\widehat{\mathbf Y}
=\operatorname{RevIN}^{-1}\!\left(
\widehat{\mathbf Y}^{\mathrm{base}}
+\overline{\boldsymbol\zeta}\odot\widehat{\mathbf E}\right),
\label{eq:finalforecast}
\end{equation}
where the gate vector is broadcast across channels. The gates share a small
initialization but are learned independently; setting them to zero recovers the
base forecast.

\begin{figure}[t]
\centering
\includegraphics[width=\linewidth]{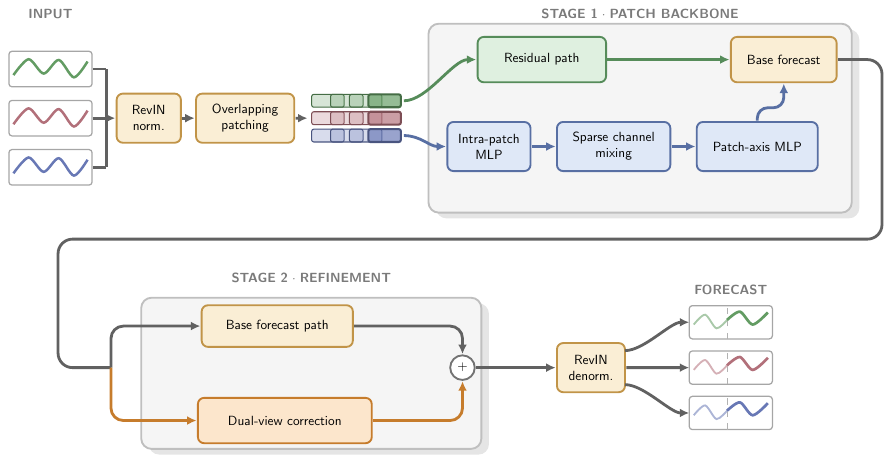}
\caption{Overview of the DuoTS architecture.}
\label{fig:lc6_architecture}
\end{figure}

Figure~\ref{fig:lc6_architecture} summarizes the complete
forecast--retrieve--revise pipeline.

The normalized window is divided into $N$ overlapping patches of length $P$
and stride $S<P$, where
$N=\lceil\max(0,L-P)/S\rceil+1$. We write
$\mathbf p_{c,n}\in\mathbb R^P$ for patch $n$ of channel $c$, ordered from
oldest to newest. Left padding, used only when needed, ensures that the final
patch ends at the forecast boundary. Residual branch weights are initialized
at or near zero; $\lambda$ denotes a tanh-transformed learned scalar, and
distinct MLP subscripts denote separate parameters unless stated otherwise.

\subsection{Two-View Correction of the Encoded History}
\label{sec:correction}
\label{sec:journey}

For each future patch, DuoTS contrasts two views of the encoded history. The
current view summarizes the tokens nearest the forecast boundary; the detail
view retrieves earlier analogs and the trajectories observed after them.
Let $\mathbf U_{c,1:N}\in\mathbb R^{N\times d}$ denote the patch embeddings
supplied by the backbone described in Section~\ref{sec:backbone}.

\begin{figure}[t]
\centering
\includegraphics[width=\linewidth]{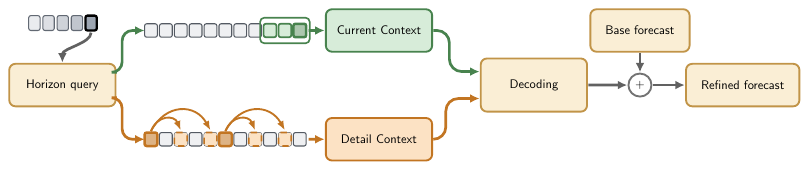}
\caption{Overview of the dual-context correction. Each future-patch query
combines recent tokens with retrieved analogs and their observed successors to
produce an additive correction.}
\label{fig:lc6_journey}
\end{figure}

\paragraph{Horizon queries and journey blocks.}
For $f=1,\ldots,F$, future patch $f$ issues a query anchored at the newest
token and offset by a learned horizon embedding
$\mathbf e_f^{\mathrm{hor}}\in\mathbb R^d$:
\begin{equation}
\boldsymbol\xi_{c,f}=\mathbf U_{c,N}+\mathbf e_f^{\mathrm{hor}}
\in\mathbb R^d.
\label{eq:horizonquery}
\end{equation}
For readability, the equations below suppress $c$; all operations are applied
separately to every channel with shared parameters. Let $B_{\mathrm{blk}}$ be
the journey-block width. The token sequence is zero-padded only on the left to
a multiple of $B_{\mathrm{blk}}$ and divided into
$M=\lceil N/B_{\mathrm{blk}}\rceil$ ordered blocks
$\mathbf G_m\in\mathbb R^{B_{\mathrm{blk}}\times d}$, $m=1,\ldots,M$. The
compressor and block-axis mixer produce
\begin{equation}
\begin{aligned}
\mathbf b_m^0
&=\operatorname{MLP}_{\mathrm{blk}}(\operatorname{vec}(\mathbf G_m))
  +\mathbf e_m^{\mathrm{blk}}\in\mathbb R^d,\\
\mathbf b_{1:M}
&=\mathbf b^0_{1:M}
  +\lambda_{\mathrm{blk}}\operatorname{MLP}_{\mathrm{baxis}}
  (\mathbf b^0_{1:M})\in\mathbb R^{M\times d},
\end{aligned}
\label{eq:journeyblocks}
\end{equation}
where $\mathbf e_m^{\mathrm{blk}}\in\mathbb R^d$ is a learned block-position
embedding. The block-axis MLP mixes the $M$ block positions independently for
each feature coordinate. The final tokens $\mathbf b_m$ form a coarse index for
detail retrieval; they are not a third correction view.

\paragraph{Cosine attention.}
We write $\operatorname{CosSim}(\mathbf x,\mathbf y)$ for the cosine
similarity between $\mathbf x$ and $\mathbf y$.
DuoTS uses $A$-head cosine cross-attention, with head width $d_a=d/A$. For a
query $\mathbf x\in\mathbb R^d$ and memory
$\mathbf Z_{1:K}=(\mathbf z_1,\ldots,\mathbf z_K)$, head $a=1,\ldots,A$ has
learned projections
$\mathbf W_Q^{(a)},\mathbf W_K^{(a)},\mathbf W_V^{(a)}
\in\mathbb R^{d_a\times d}$ and positive logit scale
$\tau_a^{\mathrm{att}}$, obtained by exponentiating a learned log-scale and
clamping the result above at $100$. Its attention weights and the complete
operator are defined as follows. Let $\alpha_j^{(a)}$ denote the
softmax-normalized weight that head $a$ assigns to memory token $j$, computed
over all $K$ memory tokens from the temperature-scaled cosine similarities
$\tau_a^{\mathrm{att}}\operatorname{CosSim}
(\mathbf W_Q^{(a)}\mathbf x,\mathbf W_K^{(a)}\mathbf z_j)$. Then
\begin{equation}
\operatorname{CosineAttention}(\mathbf x,\mathbf Z_{1:K})
=\mathbf W_O\!\left[
\sum_{j=1}^K\alpha_j^{(1)}\mathbf W_V^{(1)}\mathbf z_j
\,\middle\Vert\,\cdots\,\middle\Vert\,
\sum_{j=1}^K\alpha_j^{(A)}\mathbf W_V^{(A)}\mathbf z_j
\right]\in\mathbb R^d,
\label{eq:cosineattention}
\end{equation}
where $\mathbf W_O\in\mathbb R^{d\times d}$ is learned. Attention and output
dropout used during training are omitted from equation~\ref{eq:cosineattention}
for readability.

\paragraph{View 1: current context.}
The state at the forecast boundary is read by multi-head cosine attention from
the query to the $K_{\mathrm{cur}}$ newest tokens, where
$K_{\mathrm{cur}}$ is the current-context window size:
\begin{equation}
\mathbf h_f^{\mathrm{cur}}=\operatorname{CosineAttention}
(\boldsymbol\xi_f,\mathbf U_{N-K_{\mathrm{cur}}+1:N})
\in\mathbb R^d.
\label{eq:currentview}
\end{equation}
This context directly summarizes the dynamics closest to the forecast origin.

\paragraph{View 2: detail context.}
The detail view retrieves past states that resemble $\boldsymbol\xi_f$ and
reads what followed them. Separate learned query and key projections score
journey blocks and eligible historical patches. With
$\mathbf W_q^{x},\mathbf W_k^{x}\in\mathbb R^{d\times d}$ and positive logit
scales $\tau_x$ for $x\in\{\mathrm{blk},\mathrm{det}\}$, the scores are
\begin{equation}
\begin{aligned}
s^{\mathrm{blk}}_{f,m}
&=\tau_{\mathrm{blk}}\operatorname{CosSim}
(\mathbf W_q^{\mathrm{blk}}\boldsymbol\xi_f,
 \mathbf W_k^{\mathrm{blk}}\mathbf b_m),\\
s^{\mathrm{det}}_{f,n}
&=\tau_{\mathrm{det}}\operatorname{CosSim}
(\mathbf W_q^{\mathrm{det}}\boldsymbol\xi_f,
 \mathbf W_k^{\mathrm{det}}\mathbf U_n).
\end{aligned}
\label{eq:retrievalscores}
\end{equation}
Each $\tau_x$ is obtained by exponentiating a learned log-scale and
clamping the result above at $100$. The block retrieval budget
$K_{\mathrm{blk}}$ specifies how many highest-scoring valid blocks are retained,
and patches are ranked only inside those blocks. A patch is eligible only if
all $K_{\mathrm{suc}}$ requested successor patches, spaced by
$\nu=\lceil P/S\rceil$ patch positions, are observed.

Within the retained blocks, the model greedily selects
$K_{\mathrm{det}}$ high-scoring patches, where $K_{\mathrm{det}}$ is the patch
retrieval budget. The exclusion radius $\rho$ is measured in patch positions:
the selector prefers a new index more than $\rho$ positions from those already
selected. If that preference would leave too few candidates, it falls back to
the next highest-scoring valid patch. This avoids filling the evidence set with
near-duplicates created by overlapping patches while keeping the requested
number of matches whenever possible.

Let $n_f^{(k)}$ be the $k$-th selected historical index,
$k=1,\ldots,K_{\mathrm{det}}$. The retrieved value combines that state, its
ordered successor changes, and the future-patch query:
\begin{equation}
\begin{aligned}
\mathbf v_f^{(k)}=\operatorname{MLP}_{\mathrm{trans}}\!\Big[&
\mathbf U_{n_f^{(k)}}
\,\Vert\,(\mathbf U_{n_f^{(k)}+\nu}-\mathbf U_{n_f^{(k)}})
\,\Vert\,\cdots\\[-1mm]
&\Vert\,(\mathbf U_{n_f^{(k)}+K_{\mathrm{suc}}\nu}
-\mathbf U_{n_f^{(k)}})
\,\Vert\,\boldsymbol\xi_f\Big]\in\mathbb R^d.
\end{aligned}
\label{eq:transitionvalue}
\end{equation}
Successor changes transfer what happened after the match rather than the
absolute level at which the successor ended. Let $\mathcal V_f$ contain the
valid selected slots, and let $\pi_f^{(k)}$ be the softmax of their retrieval
scores after invalid slots are masked. The detail context is
\begin{equation}
\mathbf h_f^{\mathrm{det}}
=\chi_f^{\mathrm{blk}}\chi_f^{\mathrm{det}}
\sum_{k\in\mathcal V_f}\pi_f^{(k)}\mathbf v_f^{(k)}
\in\mathbb R^d.
\label{eq:detailaggregate}
\end{equation}
The factors $\chi_f^{\mathrm{blk}}$ and $\chi_f^{\mathrm{det}}$ are learned
sigmoid calibrations of the gap between the best and second-best block and
patch scores, respectively. Detail confidence is used only when a single patch
is retrieved; otherwise $\chi_f^{\mathrm{det}}=1$. If no runner-up exists, a
large positive margin gives confidence close to one.

\paragraph{Two-view fusion and correction.}
A sample- and horizon-conditioned two-way softmax blends the current and detail
contexts, after which an MLP decodes one correction patch:
\begin{equation}
\begin{aligned}
\boldsymbol\omega_f&=\operatorname{softmax}\!\left(
\operatorname{MLP}_{\mathrm{view}}
([\boldsymbol\xi_f\Vert\mathbf e_f^{\mathrm{hor}}])\right)
\in\mathbb R^2,\\
\overline{\mathbf h}_f
&=\omega_{f,1}\mathbf h_f^{\mathrm{cur}}
+\omega_{f,2}\mathbf h_f^{\mathrm{det}}\in\mathbb R^d,\\
\mathbf r_f&=\operatorname{MLP}_{\mathrm{dec}}
(\overline{\mathbf h}_f)\in\mathbb R^P.
\end{aligned}
\label{eq:viewfusion}
\end{equation}
Because $\boldsymbol\xi_f$ depends on the input and both terms depend on $f$,
the view weights can change across samples and forecast locations.
Reintroducing the channel index, the normalized correction is the first $H$
components of the concatenated correction patches:
\begin{equation}
\widehat{\mathbf E}_{c,:}
=\big[\mathbf r_{c,1}\Vert\cdots\Vert\mathbf r_{c,F}\big]_{1:H}
\in\mathbb R^H.
\label{eq:assemblecorrection}
\end{equation}
There is no additional projection after concatenation.
Equation~\ref{eq:finalforecast} then applies the learned future-patch gates and
inverse RevIN. Algorithm~\ref{alg:journey} gives
the complete correction procedure.

\subsection{Base Forecast: Dual-Path Backbone}
\label{sec:backbone}

The backbone provides both the base forecast and the patch embeddings used by
the correction module above. Each patch follows a structured path and a direct
path. The structured path mixes information within each patch, across related
channels, and across historical patch positions; the direct path preserves a
short linear route from the raw patch to the forecast head.

\paragraph{Sparse cross-channel mixing.}
Each patch first passes through a residual intra-patch MLP. Learned channel
embeddings then select the $\kappa_{\mathrm{ch}}$ most related channels and use
softmax attention to form their neighborhood consensus. DuoTS passes only the
difference between this consensus and the current channel through a shared MLP
and adds the result residually with a learned coefficient
$\gamma_{\mathrm{ch}}$, producing $\overline{\mathbf q}_{c,n}$. The graph is
shared across samples and patch positions; setting
$\kappa_{\mathrm{ch}}=0$ bypasses the update. Appendix~\ref{app:channel-mixer}
gives the complete construction.

\paragraph{Local and full-window patch dynamics.}
Let $d$ be the shared token width and $\Lambda$ the number of structured-path
layers. After linearly projecting $\overline{\mathbf q}_{c,n}$ to width $d$ and
adding a learned patch-position embedding, let
$\mathbf V^{(0)}_{c,n}\in\mathbb R^d$ denote the structured token. At each
layer $\ell=1,\ldots,\Lambda$, DuoTS makes first- and second-order changes
explicit:
\begin{equation}
\boldsymbol\Delta^{1,(\ell)}_{c,n}
=\mathbf V^{(\ell-1)}_{c,n}-\mathbf V^{(\ell-1)}_{c,n-1},
\qquad
\boldsymbol\Delta^{2,(\ell)}_{c,n}
=\mathbf V^{(\ell-1)}_{c,n}-2\mathbf V^{(\ell-1)}_{c,n-1}
+\mathbf V^{(\ell-1)}_{c,n-2}.
\label{eq:localdelta}
\end{equation}
Unavailable differences at the first or second patch are set to zero. Learned
scalars admit the local trend and curvature signals,
\begin{equation}
\widetilde{\mathbf V}^{(\ell)}_{c,n}
=\mathbf V^{(\ell-1)}_{c,n}
+\lambda^{1}_{\ell}\boldsymbol\Delta^{1,(\ell)}_{c,n}
+\lambda^{2}_{\ell}\boldsymbol\Delta^{2,(\ell)}_{c,n},
\label{eq:localupdate}
\end{equation}
after which a residual MLP mixes all $N$ patch positions:
\begin{equation}
\mathbf V^{(\ell)}=\widetilde{\mathbf V}^{(\ell)}
+\lambda^{\mathrm{axis}}_\ell\operatorname{MLP}_{\mathrm{axis},\ell}
(\widetilde{\mathbf V}^{(\ell)}).
\label{eq:patchaxis}
\end{equation}
Here $\mathbf V^{(\ell)},\widetilde{\mathbf V}^{(\ell)}
\in\mathbb R^{C\times N\times d}$ stack the tokens for one sample, and
$d_{\mathrm{ff}}$ is the hidden width of the axis mixer. The mixer applies a
shared $N\!\rightarrow d_{\mathrm{ff}}\!\rightarrow N$ MLP independently to
every channel and feature coordinate. Thus the differences supply
immediate-neighbor dynamics while the axis MLP allows any historical patch to
influence any other.

\paragraph{Path fusion and base forecast.}
The direct token is
$\mathbf D_{c,n}=\mathbf W_{\mathrm{dir}}\mathbf p_{c,n}+
\mathbf e^{\mathrm{pos}}_n\in\mathbb R^d$, where
$\mathbf W_{\mathrm{dir}}\in\mathbb R^{d\times P}$ and
$\mathbf e^{\mathrm{pos}}_n\in\mathbb R^d$. The same position embedding was
already added to the structured token. The effective scalar
$\lambda_{\mathrm{fuse}}$ merges the final structured token with the direct
route:
\begin{equation}
\begin{aligned}
\mathbf U_{c,n}&=\mathbf D_{c,n}+\lambda_{\mathrm{fuse}}\left(
\operatorname{MLP}_{\mathrm{fuse}}
([\mathbf V^{(\Lambda)}_{c,n}\Vert\mathbf D_{c,n}])-\mathbf D_{c,n}\right)
\in\mathbb R^d,\\
\widehat{\mathbf Y}^{\mathrm{base}}_{c,:}
&=\mathbf W_{\mathrm{head}}\operatorname{vec}
(\mathbf U_{c,1:N})+\mathbf b_{\mathrm{head}}.
\end{aligned}
\label{eq:baseforecast}
\end{equation}
Here $[\cdot\Vert\cdot]$ denotes feature concatenation,
$\operatorname{vec}$ flattens the $N$ tokens in chronological order,
$\mathbf W_{\mathrm{head}}\in\mathbb R^{H\times Nd}$, and
$\mathbf b_{\mathrm{head}}\in\mathbb R^H$. The learned fusion weight
$\lambda_{\mathrm{fuse}}$ is initialized at zero, so the backbone initially
follows the direct linear patch route. The shared flattening head maps the
fused tokens to all $H$ forecast steps in normalized coordinates. The same
tokens are also read by the correction module above.

\section{Experiments}
\label{sec:experiments}

\paragraph{Evaluation setup and protocol.}
We evaluate DuoTS on eight standard multivariate benchmarks: ETTh1, ETTh2,
ETTm1, and ETTm2 \citep{informer}, Weather \citep{weather}, Electricity
\citep{electricity}, Traffic \citep{pems}, and Solar \citep{solar}. Forecast
horizons are $H\in\{96,192,336,720\}$. The look-back length for each setting is
selected from $\{96,336,512\}$ following \citet{srsnet}. Using one window length
for all settings would not be neutral because it affects the cross-channel
structure available to a forecasting model \citep{channelindep}. Training
minimizes mean squared error, and we report MSE and MAE on standardized values.
Results are reported as the mean and sample standard deviation over three runs.
We compare against six published long-term forecasting baselines: SRSNet
\citep{srsnet}, TimeKAN \citep{timekan}, Amplifier \citep{amplifier},
iTransformer \citep{itransformer}, TimeMixer \citep{timemixer}, and PatchTST
\citep{patchtst}.
Appendices~\ref{app:repro}
and~\ref{app:tune} provide the benchmark statistics, metric definitions, and
search space.

\subsection{Main Results}
\label{sec:main-results}

\begin{table}[!t]
\centering
\caption{Long-term forecasting results for DuoTS at the horizons $H\in\{96,192,336,720\}$ with the horizon average of each dataset in the rows marked \textit{Avg}. The best MSE and MAE per row are in \textbf{bold} and the second best are \underline{underlined}.}
\label{tab:lc6-full-results}
\scriptsize
\renewcommand{\arraystretch}{0.96}
\resizebox{\textwidth}{!}{%
\begin{tabular}{llcccccccccccccc}
\toprule
 &  & \multicolumn{2}{c}{DuoTS (Ours)} & \multicolumn{2}{c}{SRSNet} & \multicolumn{2}{c}{TimeKAN} & \multicolumn{2}{c}{Amplifier} & \multicolumn{2}{c}{iTransformer} & \multicolumn{2}{c}{TimeMixer} & \multicolumn{2}{c}{PatchTST} \\
\cmidrule(lr){3-4} \cmidrule(lr){5-6} \cmidrule(lr){7-8} \cmidrule(lr){9-10} \cmidrule(lr){11-12} \cmidrule(lr){13-14} \cmidrule(lr){15-16}
Dataset & Horizon & MSE & MAE & MSE & MAE & MSE & MAE & MSE & MAE & MSE & MAE & MSE & MAE & MSE & MAE \\
\midrule
ETTh1 & 96 & \textbf{0.363}\,{\scriptsize$\pm0.0008$} & \textbf{0.388}\,{\scriptsize$\pm0.0003$} & \underline{0.366} & \underline{0.394} & 0.370 & 0.396 & 0.373 & 0.399 & 0.386 & 0.405 & 0.372 & 0.401 & 0.377 & 0.397 \\
 & 192 & \textbf{0.400}\,{\scriptsize$\pm0.0050$} & \textbf{0.413}\,{\scriptsize$\pm0.0033$} & \textbf{0.400} & \underline{0.415} & \underline{0.403} & 0.417 & 0.414 & 0.420 & 0.430 & 0.435 & 0.413 & 0.430 & 0.409 & 0.425 \\
 & 336 & \textbf{0.401}\,{\scriptsize$\pm0.0002$} & \textbf{0.430}\,{\scriptsize$\pm0.0004$} & 0.424 & \textbf{0.430} & \underline{0.420} & \underline{0.432} & 0.442 & 0.446 & 0.450 & 0.452 & 0.438 & 0.450 & 0.431 & 0.444 \\
 & 720 & \underline{0.441}\,{\scriptsize$\pm0.0023$} & \underline{0.457}\,{\scriptsize$\pm0.0016$} & \textbf{0.426} & \textbf{0.455} & 0.442 & 0.465 & 0.455 & 0.467 & 0.495 & 0.487 & 0.483 & 0.483 & 0.457 & 0.477 \\
\cmidrule(lr){1-16}
\textit{Avg} & & \textbf{0.401} & \textbf{0.422} & \underline{0.404} & \underline{0.423} & 0.408 & 0.427 & 0.421 & 0.433 & 0.440 & 0.444 & 0.426 & 0.441 & 0.418 & 0.435 \\
\midrule
ETTh2 & 96 & \textbf{0.263}\,{\scriptsize$\pm0.0006$} & \textbf{0.332}\,{\scriptsize$\pm0.0011$} & 0.271 & 0.338 & 0.280 & 0.343 & 0.287 & 0.349 & 0.292 & 0.347 & \underline{0.270} & 0.342 & 0.274 & \underline{0.337} \\
 & 192 & \textbf{0.312}\,{\scriptsize$\pm0.0001$} & \textbf{0.367}\,{\scriptsize$\pm0.0002$} & \underline{0.324} & \underline{0.372} & 0.329 & 0.382 & 0.348 & 0.393 & 0.348 & 0.384 & 0.349 & 0.387 & 0.348 & 0.384 \\
 & 336 & \textbf{0.314}\,{\scriptsize$\pm0.0018$} & \textbf{0.377}\,{\scriptsize$\pm0.0012$} & \underline{0.349} & \underline{0.394} & 0.370 & 0.412 & 0.383 & 0.423 & 0.372 & 0.407 & 0.367 & 0.410 & 0.377 & 0.416 \\
 & 720 & \textbf{0.383}\,{\scriptsize$\pm0.0027$} & \textbf{0.424}\,{\scriptsize$\pm0.0023$} & \underline{0.391} & \underline{0.434} & 0.420 & 0.450 & 0.407 & 0.444 & 0.424 & 0.444 & 0.401 & 0.436 & 0.406 & 0.441 \\
\cmidrule(lr){1-16}
\textit{Avg} & & \textbf{0.318} & \textbf{0.375} & \underline{0.333} & \underline{0.384} & 0.349 & 0.396 & 0.356 & 0.402 & 0.359 & 0.395 & 0.346 & 0.393 & 0.351 & 0.394 \\
\midrule
ETTm1 & 96 & \textbf{0.284}\,{\scriptsize$\pm0.0019$} & \underline{0.342}\,{\scriptsize$\pm0.0010$} & 0.288 & \textbf{0.341} & 0.290 & 0.348 & 0.292 & 0.346 & \underline{0.287} & \underline{0.342} & 0.293 & 0.345 & 0.289 & 0.343 \\
 & 192 & 0.334\,{\scriptsize$\pm0.0005$} & 0.366\,{\scriptsize$\pm0.0009$} & \textbf{0.326} & \textbf{0.364} & 0.332 & 0.368 & \underline{0.327} & \underline{0.365} & 0.331 & 0.371 & 0.335 & 0.372 & 0.329 & 0.368 \\
 & 336 & 0.364\,{\scriptsize$\pm0.0002$} & \textbf{0.383}\,{\scriptsize$\pm0.0005$} & 0.365 & 0.386 & \textbf{0.354} & 0.386 & 0.365 & 0.386 & \underline{0.358} & \underline{0.384} & 0.368 & 0.386 & 0.362 & 0.390 \\
 & 720 & 0.419\,{\scriptsize$\pm0.0020$} & 0.419\,{\scriptsize$\pm0.0013$} & 0.426 & 0.420 & \textbf{0.401} & \underline{0.417} & 0.427 & 0.419 & \underline{0.412} & \textbf{0.416} & 0.426 & \underline{0.417} & 0.416 & 0.423 \\
\cmidrule(lr){1-16}
\textit{Avg} & & 0.350 & \textbf{0.377} & 0.351 & \textbf{0.377} & \textbf{0.344} & 0.379 & 0.352 & 0.379 & \underline{0.347} & \underline{0.378} & 0.355 & 0.380 & 0.349 & 0.381 \\
\midrule
ETTm2 & 96 & 0.166\,{\scriptsize$\pm0.0003$} & \underline{0.255}\,{\scriptsize$\pm0.0004$} & \textbf{0.164} & \textbf{0.254} & \textbf{0.164} & \textbf{0.254} & \textbf{0.164} & \textbf{0.254} & 0.168 & 0.262 & \underline{0.165} & 0.256 & \underline{0.165} & \underline{0.255} \\
 & 192 & \textbf{0.220}\,{\scriptsize$\pm0.0011$} & \textbf{0.293}\,{\scriptsize$\pm0.0008$} & \textbf{0.220} & 0.296 & 0.238 & 0.300 & 0.226 & 0.300 & 0.224 & \underline{0.295} & 0.225 & 0.298 & \underline{0.221} & \textbf{0.293} \\
 & 336 & 0.278\,{\scriptsize$\pm0.0034$} & \underline{0.330}\,{\scriptsize$\pm0.0014$} & \textbf{0.271} & \textbf{0.327} & 0.278 & 0.331 & 0.276 & 0.331 & \underline{0.274} & \underline{0.330} & 0.277 & 0.332 & 0.276 & \textbf{0.327} \\
 & 720 & 0.364\,{\scriptsize$\pm0.0015$} & 0.385\,{\scriptsize$\pm0.0009$} & \textbf{0.353} & \textbf{0.380} & 0.359 & 0.387 & \underline{0.358} & 0.388 & 0.367 & 0.385 & 0.360 & 0.387 & 0.362 & \underline{0.381} \\
\cmidrule(lr){1-16}
\textit{Avg} & & 0.257 & \underline{0.316} & \textbf{0.252} & \textbf{0.314} & 0.259 & 0.318 & \underline{0.256} & 0.318 & 0.258 & 0.318 & \underline{0.256} & 0.318 & \underline{0.256} & \textbf{0.314} \\
\midrule
Weather & 96 & \textbf{0.142}\,{\scriptsize$\pm0.0017$} & \textbf{0.193}\,{\scriptsize$\pm0.0015$} & \underline{0.147} & \underline{0.198} & 0.151 & 0.205 & \underline{0.147} & 0.199 & 0.150 & 0.207 & 0.151 & \underline{0.198} & 0.150 & 0.200 \\
 & 192 & \textbf{0.188}\,{\scriptsize$\pm0.0019$} & \textbf{0.238}\,{\scriptsize$\pm0.0019$} & \underline{0.190} & 0.242 & 0.195 & 0.244 & 0.194 & 0.245 & 0.200 & 0.248 & 0.191 & 0.242 & 0.191 & \underline{0.239} \\
 & 336 & \textbf{0.237}\,{\scriptsize$\pm0.0038$} & \textbf{0.276}\,{\scriptsize$\pm0.0031$} & \underline{0.241} & 0.282 & 0.242 & 0.287 & 0.243 & 0.282 & 0.252 & 0.287 & 0.244 & 0.280 & 0.242 & \underline{0.279} \\
 & 720 & \underline{0.312}\,{\scriptsize$\pm0.0022$} & \underline{0.330}\,{\scriptsize$\pm0.0015$} & 0.325 & 0.340 & 0.317 & 0.340 & \textbf{0.310} & \textbf{0.329} & 0.320 & 0.336 & 0.316 & 0.331 & \underline{0.312} & \underline{0.330} \\
\cmidrule(lr){1-16}
\textit{Avg} & & \textbf{0.219} & \textbf{0.259} & 0.225 & 0.265 & 0.226 & 0.269 & \underline{0.223} & 0.263 & 0.230 & 0.269 & 0.225 & \underline{0.262} & \underline{0.223} & \underline{0.262} \\
\midrule
Electricity & 96 & \textbf{0.127}\,{\scriptsize$\pm0.0003$} & \textbf{0.222}\,{\scriptsize$\pm0.0004$} & \underline{0.131} & \underline{0.226} & 0.135 & 0.231 & 0.132 & 0.227 & 0.134 & 0.230 & 0.153 & 0.256 & 0.143 & 0.247 \\
 & 192 & \textbf{0.146}\,{\scriptsize$\pm0.0007$} & \textbf{0.241}\,{\scriptsize$\pm0.0007$} & \underline{0.147} & \textbf{0.241} & 0.149 & \underline{0.243} & 0.149 & \underline{0.243} & 0.154 & 0.250 & 0.168 & 0.269 & 0.158 & 0.260 \\
 & 336 & \textbf{0.162}\,{\scriptsize$\pm0.0006$} & \textbf{0.258}\,{\scriptsize$\pm0.0007$} & \underline{0.163} & \textbf{0.258} & 0.165 & \underline{0.260} & 0.167 & 0.261 & 0.169 & 0.265 & 0.189 & 0.281 & 0.168 & 0.267 \\
 & 720 & 0.206\,{\scriptsize$\pm0.0044$} & 0.296\,{\scriptsize$\pm0.0032$} & \underline{0.201} & \underline{0.291} & 0.206 & 0.297 & 0.203 & 0.292 & \textbf{0.194} & \textbf{0.288} & 0.228 & 0.320 & 0.214 & 0.307 \\
\cmidrule(lr){1-16}
\textit{Avg} & & \textbf{0.160} & \textbf{0.254} & \textbf{0.160} & \textbf{0.254} & 0.163 & 0.257 & \underline{0.162} & \underline{0.255} & \underline{0.162} & 0.258 & 0.184 & 0.281 & 0.170 & 0.270 \\
\midrule
Traffic & 96 & \textbf{0.355}\,{\scriptsize$\pm0.0007$} & \textbf{0.252}\,{\scriptsize$\pm0.0008$} & \underline{0.361} & \underline{0.254} & 0.388 & 0.269 & 0.391 & 0.277 & 0.363 & 0.265 & 0.369 & 0.256 & 0.370 & 0.262 \\
 & 192 & \textbf{0.380}\,{\scriptsize$\pm0.0006$} & \textbf{0.259}\,{\scriptsize$\pm0.0004$} & \textbf{0.380} & \underline{0.263} & 0.411 & 0.286 & 0.405 & 0.283 & \underline{0.384} & 0.273 & 0.400 & 0.271 & 0.386 & 0.269 \\
 & 336 & \textbf{0.386}\,{\scriptsize$\pm0.0016$} & \textbf{0.262}\,{\scriptsize$\pm0.0002$} & \underline{0.392} & \underline{0.270} & 0.425 & 0.284 & 0.416 & 0.290 & 0.396 & 0.277 & 0.407 & 0.272 & 0.396 & 0.275 \\
 & 720 & \textbf{0.424}\,{\scriptsize$\pm0.0032$} & \textbf{0.284}\,{\scriptsize$\pm0.0021$} & \underline{0.434} & \underline{0.293} & 0.455 & 0.302 & 0.436 & 0.312 & 0.445 & 0.308 & 0.462 & 0.316 & 0.435 & 0.295 \\
\cmidrule(lr){1-16}
\textit{Avg} & & \textbf{0.387} & \textbf{0.264} & \underline{0.391} & \underline{0.270} & 0.419 & 0.285 & 0.412 & 0.290 & 0.397 & 0.280 & 0.409 & 0.278 & 0.396 & 0.275 \\
\midrule
Solar & 96 & \textbf{0.162}\,{\scriptsize$\pm0.0016$} & \textbf{0.218}\,{\scriptsize$\pm0.0008$} & \underline{0.167} & \underline{0.222} & 0.187 & 0.255 & 0.175 & 0.237 & 0.174 & 0.229 & 0.180 & 0.233 & 0.170 & 0.234 \\
 & 192 & \textbf{0.177}\,{\scriptsize$\pm0.0014$} & \textbf{0.236}\,{\scriptsize$\pm0.0016$} & \underline{0.182} & \underline{0.237} & 0.194 & 0.265 & 0.198 & 0.259 & 0.205 & 0.270 & 0.201 & 0.259 & 0.204 & 0.302 \\
 & 336 & \textbf{0.185}\,{\scriptsize$\pm0.0019$} & \textbf{0.241}\,{\scriptsize$\pm0.0019$} & \underline{0.188} & \underline{0.245} & 0.203 & 0.264 & 0.213 & 0.259 & 0.216 & 0.282 & 0.214 & 0.272 & 0.212 & 0.293 \\
 & 720 & \underline{0.196}\,{\scriptsize$\pm0.0006$} & \textbf{0.250}\,{\scriptsize$\pm0.0032$} & \textbf{0.195} & \underline{0.251} & 0.209 & 0.269 & 0.222 & 0.369 & 0.211 & 0.260 & 0.218 & 0.278 & 0.215 & 0.307 \\
\cmidrule(lr){1-16}
\textit{Avg} & & \textbf{0.180} & \textbf{0.236} & \underline{0.183} & \underline{0.238} & 0.198 & 0.263 & 0.202 & 0.281 & 0.201 & 0.260 & 0.203 & 0.260 & 0.200 & 0.284 \\
\midrule
\multicolumn{2}{l}{\# Best/32} & \textbf{22} & \textbf{23} & \underline{9} & \underline{9} & 3 & 1 & 2 & 2 & 1 & 2 & 0 & 0 & 0 & 2 \\
\bottomrule
\end{tabular}%
}
\end{table}

Table~\ref{tab:lc6-full-results} reports the full comparison. DuoTS obtains the
lowest MAE in 23 of the 32 dataset--horizon settings and the lowest MSE in 22,
compared with 9 and 9 for SRSNet, the strongest baseline. Based on the horizon
averages, DuoTS leads both metrics on ETTh1, ETTh2, Weather, Traffic, and Solar.
It ties SRSNet on Electricity, achieves the best MAE on ETTm1, and trails on
ETTm2.

ETTh2 shows the clearest improvement. DuoTS reduces the horizon-average MSE by
$4.5\%$ and the MAE by $2.3\%$ relative to SRSNet and leads at all four
horizons. Traffic also provides a large and diverse retrieval pool through its
862 sensors, and DuoTS leads at every horizon on both metrics.

The correction is more challenging on the two quarter-hourly ETT subsets. At a
15-minute sampling interval the look-back window spans only a day or two, and
adjacent patches often provide similar candidates. DuoTS still matches the
best MAE on ETTm1 but gives up $1.7\%$ MSE to TimeKAN. On ETTm2 it trails
SRSNet by $2.0\%$ in horizon-average MSE.

\subsection{Ablation Study}
\label{sec:carat-corrected-screen}

Table~\ref{tab:ablation-mechanism} compares DuoTS with four training-time
variants over ETTh2, ETTm2, Weather, and Electricity. The variants remove the
second-order patch dynamics, sparse cross-channel mixing, or the intra-patch
mixer, or disable the patch-axis MLP, respectively. Each dataset column
averages the four forecasting horizons.

\raggedbottom
\begin{table}[H]
\centering
\caption{Component ablation results averaged over the four forecasting
horizons for each dataset. Each variant was tuned separately for this ablation
study. The best MSE and MAE in each column are in
\textbf{bold} and the second best are
\underline{underlined}. The DuoTS row exactly reproduces the corresponding
\textit{Avg} rows of Table~\ref{tab:lc6-full-results}.}
\label{tab:ablation-mechanism}
\scriptsize
\setlength{\tabcolsep}{3pt}
\renewcommand{\arraystretch}{1.05}
\begin{tabular*}{\textwidth}{@{\extracolsep{\fill}}lcccccccc@{}}
\toprule
\multirow{2}{*}{Case} & \multicolumn{2}{c}{ETTh2} & \multicolumn{2}{c}{ETTm2} & \multicolumn{2}{c}{Weather} & \multicolumn{2}{c}{Electricity} \\
\cmidrule(lr){2-3} \cmidrule(lr){4-5} \cmidrule(lr){6-7} \cmidrule(lr){8-9}
 & MSE & MAE & MSE & MAE & MSE & MAE & MSE & MAE \\
\midrule
DuoTS & \textbf{0.318} & \underline{0.375} & \underline{0.257} & \underline{0.316} & \textbf{0.219} & \textbf{0.259} & \underline{0.160} & 0.254 \\
\midrule
w/o second-order dynamics & \underline{0.319} & \textbf{0.374} & 0.259 & 0.318 & 0.227 & 0.265 & \underline{0.160} & \underline{0.253} \\
w/o channel mixing & \underline{0.319} & \underline{0.375} & \textbf{0.256} & \textbf{0.314} & 0.227 & 0.265 & 0.166 & 0.260 \\
w/o intra-patch mixer & 0.322 & 0.376 & 0.263 & 0.319 & \underline{0.224} & \underline{0.262} & 0.164 & 0.258 \\
w/o patch-axis MLP & 0.330 & 0.383 & 0.263 & 0.318 & \underline{0.224} & 0.263 & \textbf{0.158} & \textbf{0.252} \\
\bottomrule
\end{tabular*}
\end{table}

The full configuration is best on both metrics for Weather and has the lowest
MSE on ETTh2. Channel mixing is most useful on Electricity, while removing it
gives a small improvement on ETTm2. Second-order dynamics are beneficial in
Weather and ETTm2. Removing them improves MAE by $0.001$ on ETTh2 and
Electricity and ties the full model in Electricity MSE. The intra-patch mixer
helps on every reported dataset. Disabling the patch-axis MLP degrades ETTh2,
ETTm2, and Weather, but improves Electricity by $0.002$ in both MSE and MAE.
Thus each component contributes overall, although none yields a uniform gain
on every metric and dataset.

\begin{table}[H]
\begin{minipage}[t]{0.46\textwidth}
\vspace{0pt}
\paragraph{Which correction contexts matter?}
Table~\ref{tab:context-masking} removes the indicated context components at
inference time. ETTm2, Solar, and Weather show that removing either individual
context worsens every horizon.
Averaged over the four horizons, removing the current context increases MSE by
$0.005$--$0.014$, while removing the retrieved detail context increases it by
$0.004$--$0.012$. Removing both contexts also degrades all three datasets, by
$0.005$--$0.019$. Thus the two views provide complementary evidence rather
than an inactive auxiliary path.
\end{minipage}\hfill
\begin{minipage}[t]{0.51\textwidth}
\vspace{0pt}
\centering
\caption{Context-component ablation on representative datasets. Entries
are $\Delta$MSE ($\times10^{-3}$), averaged over four horizons. Positive values
mean that removing the indicated component hurts.}
\label{tab:context-masking}
\scriptsize
\setlength{\tabcolsep}{3.5pt}
\renewcommand{\arraystretch}{1.05}
\resizebox{\linewidth}{!}{%
\begin{tabular}{lrrr}
\toprule
Component & ETTm2 & Solar & Weather \\
\midrule
w/o Current context & +5.25 & +8.64 & +13.82 \\
w/o Detail context & +5.89 & +4.20 & +11.77 \\
w/o Both & +5.37 & +9.46 & +18.85 \\
\bottomrule
\end{tabular}
}

\end{minipage}
\end{table}

\begin{figure}[t]
\centering
\begin{minipage}[t]{0.39\textwidth}
  \centering
  \includegraphics[height=0.23\textheight]{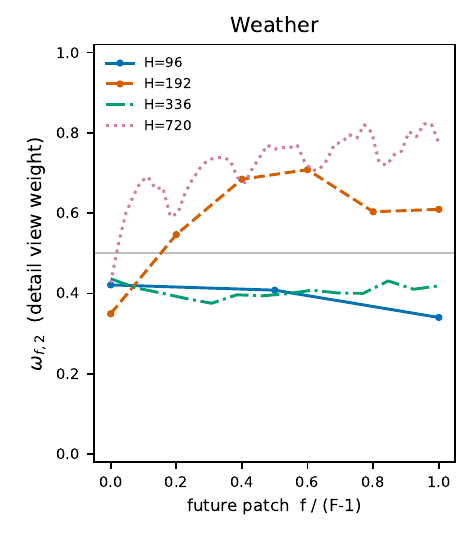}
\end{minipage}\hspace{0.025\textwidth}%
\begin{minipage}[t]{0.39\textwidth}
  \centering
  \includegraphics[height=0.23\textheight]{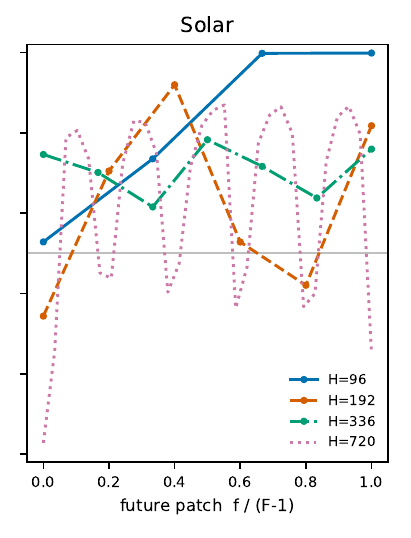}
\end{minipage}
\caption{\textbf{Full patch-wise view-weight profiles.}
Mean weight $\omega_{f,2}$ assigned to the retrieved-detail view at future
patch $f$ for Weather and Solar. The current view receives
$1-\omega_{f,2}$. Curves correspond to forecast horizons.}
\label{fig:view-weight-profiles}
\end{figure}

\paragraph{Does the view router depend on forecast position?}
Figure~\ref{fig:view-weight-profiles} shows representative full patch-wise
router trajectories for Weather and Solar. Across all 32 dataset--horizon
cells, the detail-view weight varies by at least $0.02$ in 24 cells. Among
those 24, 21 assign more detail weight farther into the forecast and three
assign less. The trajectories show both smooth horizon-dependent routing and
non-monotone, patch-specific routing, supporting an adaptive mixture across
the forecast.

\subsection{Transferring the Correction to Other Backbones}
\label{sec:module-transfer}

We add the Dual-Context Correction from DuoTS to SRSNet, TimeKAN, and PatchTST.
Across the 24 backbone--dataset--horizon cells in
Table~\ref{tab:module-transfer}, the correction lowers MSE in 15. The transfer
is not uniform. On ETTm1, adding the correction to the strong SRSNet backbone
reduces MSE from $0.426$ to $0.423$ at $H=720$. However, only two of the six
ETTm1 cells improve, while four become worse. The 15-minute sampling interval
may help explain this result. With $L=512$, adjacent patches densely cover
roughly 5.3 days and may yield highly redundant retrieved contexts. This
interpretation remains a hypothesis rather than causal evidence. Full DuoTS has
the lowest MSE in four of the eight dataset--horizon settings.

\begin{table}[H]
\centering
\caption{Correction transfer. Within each arrow, the lower endpoint is bold.
The DuoTS column is from Table~\ref{tab:lc6-full-results} and is bold when it
is lower than all three correction-equipped backbones. The second-lowest value
among the three correction-equipped endpoints and Full DuoTS is underlined.}
\label{tab:module-transfer}
\small
\setlength{\tabcolsep}{4pt}
\renewcommand{\arraystretch}{1.05}
\begin{tabular*}{\linewidth}{@{\extracolsep{\fill}}llcccc@{}}
\toprule
& & \multicolumn{3}{c}{Original $\rightarrow$ +Dual-Context Correction from DuoTS} & \\
\cmidrule(lr){3-5}
Dataset & $H$ & SRSNet & TimeKAN & PatchTST & Full DuoTS \\
\midrule
ETTh2 & 96  & 0.271 $\rightarrow$ \textbf{0.270} & 0.280 $\rightarrow$ \underline{\textbf{0.267}} & \textbf{0.274} $\rightarrow$ 0.275 & \textbf{0.263} \\
      & 720 & 0.391 $\rightarrow$ \textbf{0.382} & 0.420 $\rightarrow$ \textbf{0.409} & 0.406 $\rightarrow$ \textbf{0.393} & \underline{0.383} \\
\midrule
Weather & 96  & 0.147 $\rightarrow$ \textbf{0.145} & 0.151 $\rightarrow$ \textbf{0.141} & 0.150 $\rightarrow$ \textbf{0.143} & \underline{0.142} \\
        & 720 & 0.325 $\rightarrow$ \underline{\textbf{0.314}} & 0.317 $\rightarrow$ \textbf{0.315} & \textbf{0.312} $\rightarrow$ 0.315 & \textbf{0.312} \\
\midrule
Electricity & 96  & 0.131 $\rightarrow$ \textbf{0.130} & \textbf{0.135} $\rightarrow$ 0.135 & 0.143 $\rightarrow$ \underline{\textbf{0.129}} & \textbf{0.127} \\
            & 720 & \textbf{0.201} $\rightarrow$ \underline{0.206} & \textbf{0.206} $\rightarrow$ 0.215 & 0.214 $\rightarrow$ \textbf{0.205} & \underline{0.206} \\
\midrule
ETTm1 & 96  & \textbf{0.288} $\rightarrow$ 0.295 & 0.290 $\rightarrow$ \underline{\textbf{0.287}} & \textbf{0.289} $\rightarrow$ 0.296 & \textbf{0.284} \\
      & 720 & 0.426 $\rightarrow$ \textbf{0.423} & \textbf{0.401} $\rightarrow$ 0.405 & \textbf{0.416} $\rightarrow$ 0.425 & \underline{0.419} \\
\bottomrule
\end{tabular*}

\end{table}
\FloatBarrier

\section{Conclusion}
\label{sec:conclusion}

DuoTS combines a parallel patch backbone with a horizon-specific dual-context
correction. Each future patch balances recent information with retrieved
analogs and their observed successors to refine the base forecast. Across
eight long-term benchmarks, DuoTS obtains the lowest MSE in 22 and the lowest
MAE in 23 of 32 dataset--horizon settings. The ablation results show that the
second-order dynamics, channel mixing, and intra-patch mixing are generally
useful, while context-component ablation confirms that both correction views
affect the forecast. Gains across SRSNet, TimeKAN, and PatchTST further support
the correction as a model-agnostic module for forecasting architectures.

\paragraph{Limitations \& Future Work.}
DuoTS is limited to analogs within the finite look-back window, and weak or
duplicate matches can make the detail evidence uninformative. Its fixed-width
projections and patch-wise correction also add overhead on small-channel
series, while selected lags represent latent evidence rather than causal local
analogs. Future work will add diversity-aware retrieval and abstention, extend
memory beyond the input window, and develop lighter corrections with more
interpretable local matches and successor tests.

\subsection*{AI use statement}

Generative AI tools were used to draft portions of the manuscript, revise and
polish the writing, and assist with the retrieval and discovery of related
work. The authors checked the suggested literature against the original
sources and manually reviewed and revised all AI-assisted text. Generative AI
tools were not used to generate datasets, develop the model or its theoretical
claims, design the experiments, implement the method, preprocess the data, or
interpret the results. The authors take responsibility for the final content
of the paper, including all text, claims, citations, and artifacts produced
with the aid of generative AI.

\subsection*{Reproducibility statement}

The DuoTS architecture and its implementation-level computation are specified
in Section~\ref{sec:method}; Appendix~\ref{app:pseudocode} gives pseudocode for
the complete dual-context correction. Section~\ref{sec:experiments} reports
the experiments, while Appendices~\ref{app:repro} and~\ref{app:tune} document
their setup. The anonymous GitHub repository supplied with the submission
contains the code and configurations needed to reproduce the reported results.

\bibliography{iclr2027_conference}
\bibliographystyle{iclr2027_conference}

\appendix
\clearpage
\section{Correction Procedure in Pseudocode}
\label{app:pseudocode}

Algorithm~\ref{alg:journey} writes the implementation as a loop over future
patches while remaining vectorized over channels. A block is valid if it
contains at least one historical patch whose requested successors are all
observed. $\operatorname{TopKValid}(s,K)$ returns the indices of the
$\min(K,\text{number of valid blocks})$ highest-scoring valid blocks. Masked
blocks are never returned. $\operatorname{Eligible}(\mathcal J,K_{\mathrm{suc}},
\nu)$ then returns the patch indices assigned to blocks in $\mathcal J$ that
satisfy $n+\ell\nu\leq N$ for every
$\ell=1,\ldots,K_{\mathrm{suc}}$.

$\operatorname{Top2Margin}$ ignores masked scores and returns the best valid
score minus the second-best valid score. If only one valid score exists, it
uses a large positive sentinel margin, so the sole valid match is not compared
with a masked entry. $\operatorname{Conf}(m;\varphi,\psi)$ is its learned
sigmoid calibration, $\sigma(\exp(\varphi)m+\psi)$.
$\operatorname{DiverseTopK}_{\rho}$ greedily selects the highest-scoring
remaining candidate more than $\rho$ patch positions from every previous
selection. If none satisfies that constraint, it selects the highest-scoring
remaining valid candidate. Thus it never chooses a masked candidate or a
duplicate when enough valid candidates exist. $\operatorname{MaskedSoftmax}$
assigns zero weight to any unavailable selected slot. Finally,
$\operatorname{LeftPad}$ and $\operatorname{Block}$ create the consecutive
width-$B_{\mathrm{blk}}$ blocks, $\operatorname{TransitionValue}$ is defined in
equation~\ref{eq:transitionvalue}, and $\operatorname{Truncate}_H$ keeps the
first $H$ entries after the correction patches are concatenated in future
order.

\begin{algorithm}[H]
\caption{Dual-context analog correction for one sample}
\label{alg:journey}
\begin{algorithmic}[1]
\REQUIRE tokens $\mathbf U_{c,1:N}$ for $c=1,\ldots,C$, horizon embeddings
$\mathbf e^{\mathrm{hor}}_{1:F}$
\ENSURE normalized correction $\widehat{\mathbf E}\in\mathbb R^{C\times H}$
\STATE $\nu\leftarrow\lceil P/S\rceil$
\STATE $\mathbf G_{1:M}\leftarrow
\operatorname{Block}(\operatorname{LeftPad}(\mathbf U_{1:N}),B_{\mathrm{blk}})$
\STATE $\mathbf b^0_m\leftarrow
\operatorname{MLP}_{\mathrm{blk}}(\operatorname{vec}(\mathbf G_m))
+\mathbf e^{\mathrm{blk}}_m$ for $m=1,\ldots,M$
\STATE $\mathbf b_{1:M}\leftarrow\mathbf b^0_{1:M}
+\lambda_{\mathrm{blk}}\operatorname{MLP}_{\mathrm{baxis}}(\mathbf b^0_{1:M})$
\FOR{$f=1$ \TO $F$}
  \STATE $\boldsymbol\xi_f\leftarrow\mathbf U_N+\mathbf e_f^{\mathrm{hor}}$
  \STATE $\mathbf h_f^{\mathrm{cur}}\leftarrow
  \operatorname{CosineAttention}(\boldsymbol\xi_f,
  \mathbf U_{N-K_{\mathrm{cur}}+1:N})$
  \STATE $s^{\mathrm{blk}}_{f,m}\leftarrow
  \tau_{\mathrm{blk}}\operatorname{CosSim}
  (\mathbf W_q^{\mathrm{blk}}\boldsymbol\xi_f,
  \mathbf W_k^{\mathrm{blk}}\mathbf b_m)$ for $m=1,\ldots,M$
  \STATE $\mathcal J_f\leftarrow
  \operatorname{TopKValid}(s^{\mathrm{blk}}_{f,1:M},K_{\mathrm{blk}})$
  \STATE $m_f^{\mathrm{blk}}\leftarrow
  \operatorname{Top2Margin}(s^{\mathrm{blk}}_{f,1:M})$
  \STATE $\chi_f^{\mathrm{blk}}\leftarrow
  \operatorname{Conf}(m_f^{\mathrm{blk}};
  \varphi_{\mathrm{blk}}^\chi,\psi_{\mathrm{blk}}^\chi)$
  \STATE $\mathcal C_f\leftarrow
  \operatorname{Eligible}(\mathcal J_f,K_{\mathrm{suc}},\nu)$
  \STATE $s^{\mathrm{det}}_{f,n}\leftarrow
  \tau_{\mathrm{det}}\operatorname{CosSim}
  (\mathbf W_q^{\mathrm{det}}\boldsymbol\xi_f,
  \mathbf W_k^{\mathrm{det}}\mathbf U_n)$ for $n\in\mathcal C_f$
  \STATE $\mathcal I_f\leftarrow
  \operatorname{DiverseTopK}_{\rho}
  (s^{\mathrm{det}}_{f,\mathcal C_f},K_{\mathrm{det}})$
  \FOR{$k=1$ \TO $K_{\mathrm{det}}$}
    \STATE $n_f^{(k)}\leftarrow\mathcal I_f[k]$
    \STATE $\mathbf v_f^{(k)}\leftarrow
    \operatorname{TransitionValue}(\mathbf U,n_f^{(k)},\nu,
    K_{\mathrm{suc}},\boldsymbol\xi_f)$
  \ENDFOR
  \STATE $\boldsymbol\pi_f\leftarrow
  \operatorname{MaskedSoftmax}(s^{\mathrm{det}}_{f,\mathcal I_f})$
  \STATE $\mathbf h_f^{\mathrm{det}}\leftarrow
  \chi_f^{\mathrm{blk}}\sum_{k=1}^{K_{\mathrm{det}}}
  \pi_f^{(k)}\mathbf v_f^{(k)}$
  \IF{$K_{\mathrm{det}}=1$}
    \STATE $m_f^{\mathrm{det}}\leftarrow
    \operatorname{Top2Margin}(s^{\mathrm{det}}_{f,\mathcal C_f})$
    \STATE $\mathbf h_f^{\mathrm{det}}\leftarrow
    \operatorname{Conf}(m_f^{\mathrm{det}};
    \varphi_{\mathrm{det}}^\chi,\psi_{\mathrm{det}}^\chi)
    \mathbf h_f^{\mathrm{det}}$
  \ENDIF
  \STATE $\boldsymbol\omega_f\leftarrow\operatorname{softmax}
  (\operatorname{MLP}_{\mathrm{view}}
  ([\boldsymbol\xi_f\Vert\mathbf e_f^{\mathrm{hor}}]))$
  \STATE $\mathbf r_f\leftarrow
  \operatorname{MLP}_{\mathrm{dec}}(
  \omega_{f,1}\mathbf h_f^{\mathrm{cur}}+
  \omega_{f,2}\mathbf h_f^{\mathrm{det}})$
\ENDFOR
\STATE $\widehat{\mathbf E}\leftarrow
\operatorname{Truncate}_H(\operatorname{Concat}
(\mathbf r_1,\ldots,\mathbf r_F))$
\RETURN $\widehat{\mathbf E}$
\end{algorithmic}
\end{algorithm}

\section{Sparse Channel Mixer Details}
\label{app:channel-mixer}

The main text retains only the channel residual update. For each channel $c$,
let $\mathbf e_c^q,\mathbf e_c^k\in\mathbb R^{d_g}$ be learned query and key
embeddings of graph width $d_g$, and let $\beta\in\mathbb R$ be the learned
self-channel logit. The omitted graph construction is
\begin{equation}
\begin{aligned}
\mathbf q_{c,n}
&=\mathbf p_{c,n}+\operatorname{MLP}_{\mathrm{in}}(\mathbf p_{c,n}),\\
s_{c,j}
&=\frac{(\mathbf e_c^q)^\top\mathbf e_j^k}{\sqrt{d_g}},
&\mathcal N_c
&=\operatorname{TopK}_{j\ne c}(s_{c,j},\kappa_{\mathrm{ch}}),\\
\left(a_{c,c},(a_{c,j})_{j\in\mathcal N_c}\right)
&=\operatorname{softmax}\left(\beta,
(s_{c,j})_{j\in\mathcal N_c}\right),
&\mathbf m_{c,n}
&=\sum_{j\in\mathcal N_c\cup\{c\}}a_{c,j}\mathbf q_{j,n}.
\end{aligned}
\label{eq:channelgraph-detail}
\end{equation}
The residual coefficient $\gamma_{\mathrm{ch}}$ is also learned. The graph is
shared across samples and patch positions. Section~\ref{sec:backbone} describes
how its consensus enters the residual channel update.

\section{Reproducibility Details}
\label{app:repro}

\subsection{Benchmarks and Their Statistics}
\label{app:bench}

Table~\ref{tab:carat-benchmarks} summarizes the eight public benchmarks. The
four ETT datasets contain seven transformer load and oil-temperature signals.
ETTh1 and ETTh2 are sampled hourly, while ETTm1 and ETTm2 use 15-minute
intervals \citep{informer}. Weather contains 21 meteorological variables
recorded every ten minutes \citep{weather}. Electricity reports hourly demand
from 321 clients \citep{electricity}, Traffic contains hourly occupancy from
862 freeway sensors \citep{pems}, and Solar records ten-minute generation from
137 photovoltaic plants \citep{solar}.

\begin{table}[t]
\centering
\caption{Statistics of the eight long-term forecasting benchmarks. $C$ is the
number of channels and $T$ is the number of observations. Dataset partitions
follow chronological order.}
\label{tab:carat-benchmarks}
\small
\setlength{\tabcolsep}{4pt}
\renewcommand{\arraystretch}{0.96}
\resizebox{\columnwidth}{!}{%
\begin{tabular}{llrrrl}
\toprule
Benchmark & Recorded quantity & $C$ & $T$ & Interval & Train / val / test \\
\midrule
ETTh1 \citep{informer} & Transformer load and oil temperature & 7 & 17,420 & 1 h & 8,640 / 2,880 / 2,880 \\
ETTh2 \citep{informer} & Transformer load and oil temperature & 7 & 17,420 & 1 h & 8,640 / 2,880 / 2,880 \\
ETTm1 \citep{informer} & Transformer load and oil temperature & 7 & 69,680 & 15 min & 34,560 / 11,520 / 11,520 \\
ETTm2 \citep{informer} & Transformer load and oil temperature & 7 & 69,680 & 15 min & 34,560 / 11,520 / 11,520 \\
Weather \citep{weather} & Meteorological indicators & 21 & 52,696 & 10 min & 36,887 / 5,270 / 10,539 \\
Electricity \citep{electricity} & Client-level power demand & 321 & 26,304 & 1 h & 18,412 / 2,632 / 5,260 \\
Traffic \citep{pems} & Freeway sensor occupancy & 862 & 17,544 & 1 h & 12,280 / 1,756 / 3,508 \\
Solar \citep{solar} & Photovoltaic plant output & 137 & 52,560 & 10 min & 36,792 / 5,256 / 10,512 \\
\bottomrule
\end{tabular}%
}
\end{table}

\subsection{Reported Error Measures}
\label{app:measures}

Let $i$ index the $n$ test windows, $h$ the forecast steps, and $c$ the
channels. We report
\begin{equation}
\label{eq:carat-metrics}
\mathrm{MSE}=\frac{1}{nHC}\sum_{i,h,c}
\big(\widehat Y_{i,h,c}-Y_{i,h,c}\big)^2,
\qquad
\mathrm{MAE}=\frac{1}{nHC}\sum_{i,h,c}
\big|\widehat Y_{i,h,c}-Y_{i,h,c}\big|.
\end{equation}
Every window, forecast step, and channel receives equal weight. MSE is used as
the training objective and gives greater weight to large errors. MAE provides a
complementary measure that is less sensitive to large individual residuals.
Lower values are better for both metrics.

\section{Evaluation Setup Details}
\label{app:tune}

\subsection{Procedure and Search Space}
\label{app:space}

We use Optuna \citep{optuna} with the Tree-structured Parzen Estimator, a
multivariate group sampler, 15 startup trials, and a fixed sampler seed. A
separate study is run for each benchmark, look-back candidate, and forecast
horizon. Each study minimizes the best validation MSE reached during training.
Most studies use 50 trials, while a small number use 20 to 30 trials. The
configuration with the lowest validation MSE is evaluated on the test split.
Table~\ref{tab:carat-space} lists the complete search space.

\begin{table}[t]
\centering
\caption{Unified hyperparameter search space for DuoTS. Derived values are
computed from the sampled coordinates, and retrieval ranges are capped by the
available patch grid.}
\label{tab:carat-space}
\small
\setlength{\tabcolsep}{4pt}
\renewcommand{\arraystretch}{0.96}
\resizebox{\columnwidth}{!}{%
\begin{tabular}{lll}
\toprule
Coordinate & Symbol & Range, cap, or fixed value \\
\midrule
\multicolumn{3}{l}{\textit{Sampled}} \\
Learning rate & $\eta$ & log-uniform, $[2\times10^{-5},3\times10^{-3}]$ \\
Path width & $d$ & $\{32,64,128,256,384,512\}$ \\
Mixer hidden width & $d_{\mathrm{ff}}$ & $\{16,32,64,128,256,512,1024\}$ \\
Structured-path layers & $\Lambda$ & $\{1,2,3\}$ \\
Dropout & & uniform, $[0,0.40]$ \\
Patch length & $P$ & $\{12,24,32,48,64,96,112,128,144,168,192,256\}$, capped at $L$ \\
Stride ratio & $r$ & $\{0.30,0.50,0.75\}$ \\
Journey block width & $B_{\mathrm{blk}}$ & $\{2,3,4,6,8\}$, capped at $N$ \\
Current-context window & $K_{\mathrm{cur}}$ & $\{2,3,4,6\}$, capped at $N$ \\
Successor span & $K_{\mathrm{suc}}$ & integer in $[1,4]$, capped at $\lfloor (N-1)/\nu\rfloor$ \\
Block retrieval budget & $K_{\mathrm{blk}}$ & $\{1,2\}$ \\
Patch retrieval budget & $K_{\mathrm{det}}$ & $\{1,2,3\}$ \\
Exclusion radius & $\rho$ & $\{1,2,3,4\}$, sampled only when $K_{\mathrm{det}}>1$ \\
Channel neighbourhood & $\kappa_{\mathrm{ch}}$ & $\{0,1,2,4,6,8,12,16,20,32,64,128\}$, capped at $C-1$ \\
Channel graph width & $d_g$ & $\{4,8,16,32,64\}$ \\
Initial channel gate & & log-uniform, $[10^{-4},5\times10^{-3}]$ \\
Initial self logit & $\beta$ & uniform, $[1,4]$ \\
Initial correction gate & & $\{0,0.01,0.03,0.05\}$ \\
Initial patch-axis gate & & $\{0.025,0.05,0.10,0.20\}$ \\
\midrule
\multicolumn{3}{l}{\textit{Derived or fixed}} \\
Stride & $S$ & $\max(1,\mathrm{round}(rP))$, derived \\
Patch count & $N$ & $\lceil\max(0,L-P)/S\rceil+1$, derived \\
Successor spacing & $\nu$ & $\lceil P/S\rceil$, derived \\
Correction heads & & fixed at 4 \\
Fusion-gate initialization & $\lambda_{\mathrm{fuse}}$ & fixed at 0 and learned thereafter \\
Weight decay & & fixed at 0 \\
Normalization & & fixed, RevIN without a learned affine map \\
\bottomrule
\end{tabular}%
}
\end{table}

Two constraints keep sampled configurations feasible. A trial is pruned when
its requested successor span or current-context window exceeds the available
patch grid. The exclusion radius is sampled only when more than one detail
patch is retrieved because a single match does not require a diversity
constraint.

\subsection{Selected Configurations}
\label{app:selected}

Tables~\ref{tab:carat-cfg-backbone} and~\ref{tab:carat-cfg-correction} report
the best configuration for every benchmark and forecast horizon.

\begin{table*}[t]
\centering
\caption{Backbone settings of the reported DuoTS models. The look-back length $L$ is chosen per cell from $\{96,336,512\}$, $P$ and $S$ are the patch length and stride, $d$ is the shared path width, $d_{\mathrm{ff}}$ the hidden width of the mixer MLPs, and $\Lambda$ the number of structured-path layers. Dropout is shared by the backbone MLPs, the channel mixer, and the correction attention. $\eta$ is the Adam learning rate and $B$ the batch size.}
\label{tab:carat-cfg-backbone}
\resizebox{\textwidth}{!}{%
\begin{tabular}{llrrrrrrrrr}
\toprule
Benchmark & $H$ & $L$ & $P$ & $S$ & $d$ & $d_{\mathrm{ff}}$ & $\Lambda$ & Dropout & $\eta$ & $B$ \\
\midrule
ETTh1 & 96 & 336 & 16 & 8 & 512 & 256 & 2 & 0.274 & $2.585\times 10^{-4}$ & 32 \\
ETTh1 & 192 & 512 & 32 & 10 & 384 & 64 & 2 & 0.200 & $1.601\times 10^{-4}$ & 32 \\
ETTh1 & 336 & 512 & 112 & 56 & 384 & 256 & 2 & 0.050 & $9.300\times 10^{-5}$ & 32 \\
ETTh1 & 720 & 336 & 16 & 8 & 128 & 128 & 2 & 0.256 & $2.443\times 10^{-4}$ & 32 \\
\midrule
ETTh2 & 96 & 512 & 48 & 14 & 256 & 256 & 1 & 0.200 & $2.324\times 10^{-3}$ & 256 \\
ETTh2 & 192 & 512 & 24 & 7 & 32 & 32 & 1 & 0.300 & $2.299\times 10^{-4}$ & 256 \\
ETTh2 & 336 & 512 & 48 & 24 & 128 & 512 & 2 & 0.200 & $1.720\times 10^{-4}$ & 256 \\
ETTh2 & 720 & 336 & 32 & 16 & 32 & 16 & 2 & 0.116 & $9.332\times 10^{-4}$ & 32 \\
\midrule
ETTm1 & 96 & 336 & 12 & 6 & 256 & 256 & 2 & 0.252 & $9.929\times 10^{-4}$ & 32 \\
ETTm1 & 192 & 512 & 24 & 18 & 384 & 256 & 1 & 0.300 & $5.821\times 10^{-5}$ & 128 \\
ETTm1 & 336 & 512 & 32 & 16 & 256 & 128 & 2 & 0.400 & $1.397\times 10^{-4}$ & 256 \\
ETTm1 & 720 & 336 & 16 & 8 & 64 & 1024 & 2 & 0.239 & $1.590\times 10^{-4}$ & 32 \\
\midrule
ETTm2 & 96 & 336 & 12 & 6 & 256 & 16 & 2 & 0.339 & $5.270\times 10^{-5}$ & 32 \\
ETTm2 & 192 & 512 & 12 & 6 & 64 & 128 & 2 & 0.400 & $9.969\times 10^{-5}$ & 256 \\
ETTm2 & 336 & 336 & 12 & 9 & 128 & 512 & 3 & 0.200 & $4.689\times 10^{-5}$ & 128 \\
ETTm2 & 720 & 336 & 32 & 16 & 256 & 256 & 2 & 0.001 & $1.154\times 10^{-4}$ & 32 \\
\midrule
Weather & 96 & 512 & 32 & 16 & 128 & 32 & 2 & 0.304 & $6.539\times 10^{-4}$ & 64 \\
Weather & 192 & 512 & 32 & 10 & 128 & 32 & 2 & 0.300 & $6.472\times 10^{-4}$ & 64 \\
Weather & 336 & 512 & 24 & 18 & 256 & 256 & 2 & 0.400 & $7.906\times 10^{-4}$ & 64 \\
Weather & 720 & 512 & 16 & 8 & 32 & 64 & 2 & 0.244 & $3.245\times 10^{-4}$ & 32 \\
\midrule
Electricity & 96 & 512 & 48 & 24 & 256 & 32 & 2 & 0.068 & $3.719\times 10^{-4}$ & 32 \\
Electricity & 192 & 512 & 112 & 84 & 256 & 16 & 1 & 0.300 & $1.151\times 10^{-3}$ & 16 \\
Electricity & 336 & 512 & 112 & 34 & 32 & 512 & 2 & 0.050 & $8.048\times 10^{-4}$ & 16 \\
Electricity & 720 & 512 & 16 & 8 & 32 & 128 & 2 & 0.159 & $8.650\times 10^{-4}$ & 32 \\
\midrule
Traffic & 96 & 512 & 128 & 38 & 256 & 128 & 2 & 0.200 & $1.811\times 10^{-3}$ & 16 \\
Traffic & 192 & 512 & 96 & 72 & 256 & 16 & 2 & 0.250 & $2.107\times 10^{-3}$ & 16 \\
Traffic & 336 & 512 & 48 & 24 & 64 & 512 & 2 & 0.250 & $2.706\times 10^{-3}$ & 8 \\
Traffic & 720 & 512 & 128 & 96 & 256 & 128 & 2 & 0.150 & $2.333\times 10^{-3}$ & 8 \\
\midrule
Solar & 96 & 512 & 24 & 18 & 512 & 16 & 1 & 0.050 & $4.754\times 10^{-4}$ & 16 \\
Solar & 192 & 512 & 32 & 10 & 256 & 16 & 3 & 0.100 & $5.044\times 10^{-4}$ & 16 \\
Solar & 336 & 512 & 48 & 14 & 32 & 256 & 1 & 0.400 & $6.104\times 10^{-4}$ & 16 \\
Solar & 720 & 512 & 24 & 7 & 64 & 32 & 1 & 0.000 & $5.151\times 10^{-4}$ & 16 \\
\bottomrule
\end{tabular}%
}
\end{table*}

\begin{table*}[t]
\centering
\caption{Correction and channel settings of the reported DuoTS models. $B_{\mathrm{blk}}$ is the journey block width, $K_{\mathrm{cur}}$ the number of newest tokens read by the current context, $K_{\mathrm{suc}}$ the successor span of a retrieved analog, $K_{\mathrm{blk}}$ and $K_{\mathrm{det}}$ the block and patch retrieval budgets, and $\rho$ the exclusion radius, which is inactive whenever $K_{\mathrm{det}}=1$. The channel columns give the neighbourhood size $\kappa_{\mathrm{ch}}$, the graph width $d_g$, the initial self logit $\beta$, and the initial channel gate. The final two columns give the initial correction gate and the initial patch-axis gate.}
\label{tab:carat-cfg-correction}
\resizebox{\textwidth}{!}{%
\begin{tabular}{llrrrrrrrrrrrr}
\toprule
Benchmark & $H$ & $B_{\mathrm{blk}}$ & $K_{\mathrm{cur}}$ & $K_{\mathrm{suc}}$ & $K_{\mathrm{blk}}$ & $K_{\mathrm{det}}$ & $\rho$ & $\kappa_{\mathrm{ch}}$ & $d_g$ & $\beta$ & Gate init & Corr.\ init & Patch init \\
\midrule
ETTh1 & 96 & 3 & 4 & 4 & 1 & 1 & 0 & 0 & 4 & 1.27 & $3.03\times 10^{-3}$ & 0.000 & 0.100 \\
ETTh1 & 192 & 4 & 4 & 3 & 2 & 1 & 0 & 0 & 8 & 2.23 & $8.08\times 10^{-4}$ & 0.000 & 0.200 \\
ETTh1 & 336 & 4 & 2 & 1 & 1 & 2 & 1 & 1 & 8 & 2.64 & $2.35\times 10^{-3}$ & 0.000 & 0.200 \\
ETTh1 & 720 & 6 & 3 & 3 & 2 & 1 & 0 & 0 & 4 & 3.01 & $4.90\times 10^{-3}$ & 0.000 & 0.100 \\
\midrule
ETTh2 & 96 & 4 & 6 & 3 & 1 & 1 & 0 & 2 & 4 & 1.71 & $8.45\times 10^{-4}$ & 0.010 & 0.050 \\
ETTh2 & 192 & 4 & 2 & 3 & 2 & 1 & 0 & 4 & 4 & 2.29 & $6.25\times 10^{-4}$ & 0.000 & 0.100 \\
ETTh2 & 336 & 6 & 6 & 3 & 2 & 3 & 4 & 0 & 32 & 1.22 & $1.01\times 10^{-4}$ & 0.050 & 0.100 \\
ETTh2 & 720 & 2 & 3 & 4 & 2 & 2 & 2 & 4 & 4 & 2.02 & $3.66\times 10^{-4}$ & 0.000 & 0.100 \\
\midrule
ETTm1 & 96 & 6 & 3 & 3 & 1 & 1 & 0 & 0 & 32 & 3.88 & $1.75\times 10^{-3}$ & 0.000 & 0.100 \\
ETTm1 & 192 & 2 & 6 & 2 & 2 & 1 & 0 & 2 & 8 & 2.74 & $1.67\times 10^{-4}$ & 0.050 & 0.050 \\
ETTm1 & 336 & 6 & 2 & 2 & 1 & 2 & 2 & 4 & 4 & 3.96 & $1.44\times 10^{-4}$ & 0.050 & 0.100 \\
ETTm1 & 720 & 3 & 4 & 2 & 2 & 2 & 4 & 0 & 32 & 3.87 & $2.83\times 10^{-4}$ & 0.000 & 0.100 \\
\midrule
ETTm2 & 96 & 4 & 3 & 3 & 2 & 3 & 1 & 2 & 32 & 3.67 & $3.17\times 10^{-4}$ & 0.000 & 0.100 \\
ETTm2 & 192 & 3 & 2 & 2 & 1 & 3 & 3 & 4 & 4 & 3.10 & $5.42\times 10^{-4}$ & 0.050 & 0.100 \\
ETTm2 & 336 & 8 & 3 & 4 & 1 & 3 & 3 & 4 & 8 & 1.51 & $2.06\times 10^{-4}$ & 0.050 & 0.200 \\
ETTm2 & 720 & 2 & 2 & 1 & 1 & 3 & 3 & 0 & 32 & 1.37 & $4.49\times 10^{-4}$ & 0.000 & 0.100 \\
\midrule
Weather & 96 & 6 & 2 & 3 & 1 & 2 & 3 & 16 & 64 & 2.47 & $2.79\times 10^{-3}$ & 0.000 & 0.100 \\
Weather & 192 & 2 & 4 & 4 & 2 & 2 & 4 & 12 & 4 & 2.42 & $1.99\times 10^{-4}$ & 0.010 & 0.025 \\
Weather & 336 & 4 & 3 & 2 & 2 & 1 & 0 & 12 & 32 & 2.17 & $6.73\times 10^{-4}$ & 0.010 & 0.025 \\
Weather & 720 & 3 & 2 & 3 & 1 & 1 & 0 & 8 & 8 & 1.87 & $2.05\times 10^{-3}$ & 0.000 & 0.100 \\
\midrule
Electricity & 96 & 3 & 4 & 4 & 1 & 3 & 2 & 128 & 8 & 3.64 & $1.01\times 10^{-4}$ & 0.000 & 0.100 \\
Electricity & 192 & 3 & 3 & 1 & 1 & 1 & 0 & 64 & 32 & 1.93 & $6.30\times 10^{-4}$ & 0.050 & 0.200 \\
Electricity & 336 & 3 & 3 & 2 & 2 & 2 & 3 & 0 & 4 & 1.29 & $2.19\times 10^{-4}$ & 0.030 & 0.200 \\
Electricity & 720 & 6 & 2 & 2 & 2 & 1 & 0 & 8 & 32 & 2.96 & $1.27\times 10^{-3}$ & 0.000 & 0.100 \\
\midrule
Traffic & 96 & 3 & 3 & 2 & 1 & 2 & 3 & 2 & 32 & 1.06 & $1.51\times 10^{-3}$ & 0.030 & 0.100 \\
Traffic & 192 & 6 & 2 & 1 & 2 & 3 & 4 & 8 & 16 & 1.88 & $2.22\times 10^{-4}$ & 0.010 & 0.025 \\
Traffic & 336 & 2 & 6 & 4 & 2 & 3 & 2 & 4 & 16 & 2.72 & $2.75\times 10^{-4}$ & 0.010 & 0.050 \\
Traffic & 720 & 4 & 4 & 2 & 1 & 1 & 0 & 2 & 16 & 3.24 & $3.56\times 10^{-3}$ & 0.010 & 0.100 \\
\midrule
Solar & 96 & 2 & 3 & 2 & 2 & 1 & 0 & 16 & 16 & 3.21 & $1.66\times 10^{-4}$ & 0.010 & 0.100 \\
Solar & 192 & 8 & 3 & 4 & 2 & 1 & 0 & 16 & 64 & 1.08 & $2.31\times 10^{-3}$ & 0.000 & 0.025 \\
Solar & 336 & 6 & 3 & 3 & 1 & 3 & 4 & 8 & 16 & 2.11 & $4.36\times 10^{-4}$ & 0.030 & 0.050 \\
Solar & 720 & 6 & 2 & 4 & 2 & 2 & 3 & 16 & 32 & 1.57 & $1.01\times 10^{-3}$ & 0.050 & 0.050 \\
\bottomrule
\end{tabular}%
}
\end{table*}

\FloatBarrier

\subsection{Meaning of Each Searched Coordinate}
\label{app:meaning}

\paragraph{Path width $d$.}
This shared width is used by the direct path, structured path, horizon queries,
journey tokens, and both correction contexts. Increasing $d$ therefore expands
both the backbone and the correction module.

\paragraph{Mixer hidden width $d_{\mathrm{ff}}$.}
The hidden width controls the intra-patch MLP, patch-axis mixer, path fusion,
and correction decoder. It changes their capacity without changing the stored
token width.

\paragraph{Structured-path layers $\Lambda$.}
This value specifies how often the structured path applies local difference
features and full-axis mixing. It does not change the depth of the direct path.

\paragraph{Patch length and stride ratio.}
Together they determine the number of historical patches $N$. More overlap
creates a larger pool with more similar candidates, while less overlap creates
a smaller pool whose patches are farther apart.

\paragraph{Journey block width $B_{\mathrm{blk}}$.}
The block width specifies how many consecutive patches form one journey token.
It controls the resolution of coarse retrieval and the number of journey
blocks $M=\lceil N/B_{\mathrm{blk}}\rceil$.

\paragraph{Retrieval budgets $K_{\mathrm{blk}}$, $K_{\mathrm{det}}$, and the
radius $\rho$.}
$K_{\mathrm{blk}}$ controls how many coarse blocks continue to patch-level
retrieval. $K_{\mathrm{det}}$ determines the number of detail matches. The
radius discourages selected matches from occupying nearby patch positions when
more separated candidates remain available.

\paragraph{Current-context window $K_{\mathrm{cur}}$ and successor span
$K_{\mathrm{suc}}$.}
$K_{\mathrm{cur}}$ sets the number of newest tokens read by the current
context. $K_{\mathrm{suc}}$ determines how many observed successor patches are
attached to each retrieved match. A longer successor span provides more
continuation evidence but reduces the number of eligible historical patches.

\paragraph{Channel neighbourhood $\kappa_{\mathrm{ch}}$, graph width $d_g$,
and self logit $\beta$.}
$\kappa_{\mathrm{ch}}$ is the number of other channels used by the sparse
channel graph. Setting it to zero disables cross-channel mixing. The graph
width controls the channel query and key embeddings, while $\beta$ controls the
self-edge before the channel weights are normalized.

\paragraph{Gate initializations.}
The search includes the initial channel weight, correction scale, and patch-axis
weight. The underlying parameter of $\lambda_{\mathrm{fuse}}$ in
equation~\ref{eq:baseforecast} is always initialized at zero and learned
thereafter. Its initialization is not searched. These small initial values allow
the additional paths to enter the forecast gradually during optimization.

\section{Comparison with Retrieval-Augmented Forecasters}
\label{app:retrieval-forecasters}

Table~\ref{tab:retrieval-paper-comparison} collects the complete reported
results of RAFT and SARAF on the seven datasets shared with DuoTS. The table
retains every forecast horizon and the dataset averages. The source papers use
a look-back of 720 whereas DuoTS selects its look-back per cell. Every selected
DuoTS look-back is shorter at 96, 336, or 512. Despite using less historical
context, DuoTS has the lowest reported MSE in 18 of the 28 dataset--horizon
rows and the lowest or tied-lowest MAE in 22. These values are still a
contextual cross-paper comparison rather than a controlled rerun.

\begin{table}[!t]
\centering
\caption{Reported full-horizon comparison with retrieval-augmented forecasters (lower is better). DuoTS uses the per-cell look-back selected from $\{96,336,512\}$. RAFT and SARAF use $L=720$. RAFT values are from Tables~16--17 of \citet{raft}, and SARAF values are from Table~4 of \citet{saraf}. These are cross-paper reported values rather than a controlled rerun. Within the three displayed methods, the best value is bold and the second-best is underlined.}
\label{tab:retrieval-paper-comparison}
\scriptsize
\renewcommand{\arraystretch}{0.94}
\setlength{\tabcolsep}{3.2pt}
\begin{tabular}{llcccccc}
\toprule
 & & \multicolumn{2}{c}{DuoTS} & \multicolumn{2}{c}{RAFT} & \multicolumn{2}{c}{SARAF} \\
\cmidrule(lr){3-4}\cmidrule(lr){5-6}\cmidrule(lr){7-8}
Dataset & $H$ & MSE & MAE & MSE & MAE & MSE & MAE \\
\midrule
ETTh1 & 96 & \textbf{0.363} & \textbf{0.388} & \underline{0.367} & \underline{0.397} & 0.371 & 0.399 \\
 & 192 & \textbf{0.400} & \textbf{0.413} & 0.411 & 0.427 & \underline{0.404} & \underline{0.420} \\
 & 336 & \textbf{0.401} & \textbf{0.430} & 0.436 & \underline{0.442} & \underline{0.435} & 0.447 \\
 & 720 & \textbf{0.441} & \textbf{0.457} & 0.467 & 0.478 & \underline{0.451} & \underline{0.473} \\
\cmidrule(lr){2-8}
 & \textit{Avg} & \textbf{0.401} & \textbf{0.422} & 0.420 & 0.436 & \underline{0.415} & \underline{0.435} \\
\midrule
ETTh2 & 96 & \textbf{0.263} & \textbf{0.332} & 0.276 & 0.344 & \underline{0.271} & \underline{0.335} \\
 & 192 & \textbf{0.312} & \textbf{0.367} & 0.347 & 0.393 & \underline{0.331} & \underline{0.378} \\
 & 336 & \textbf{0.314} & \textbf{0.377} & 0.376 & 0.425 & \underline{0.359} & \underline{0.408} \\
 & 720 & \textbf{0.383} & \textbf{0.424} & 0.436 & 0.473 & \underline{0.399} & \underline{0.447} \\
\cmidrule(lr){2-8}
 & \textit{Avg} & \textbf{0.318} & \textbf{0.375} & 0.359 & 0.409 & \underline{0.340} & \underline{0.392} \\
\midrule
ETTm1 & 96 & \textbf{0.284} & \textbf{0.342} & 0.302 & 0.349 & \underline{0.297} & \underline{0.348} \\
 & 192 & 0.334 & \textbf{0.366} & \textbf{0.329} & \underline{0.367} & \underline{0.331} & 0.370 \\
 & 336 & 0.364 & \textbf{0.383} & \textbf{0.355} & \textbf{0.383} & \underline{0.357} & \underline{0.386} \\
 & 720 & 0.419 & 0.419 & \underline{0.406} & \underline{0.413} & \textbf{0.401} & \textbf{0.410} \\
\cmidrule(lr){2-8}
 & \textit{Avg} & 0.350 & \textbf{0.377} & \underline{0.348} & \underline{0.378} & \textbf{0.346} & \underline{0.378} \\
\midrule
ETTm2 & 96 & 0.166 & \underline{0.255} & \underline{0.164} & 0.256 & \textbf{0.160} & \textbf{0.252} \\
 & 192 & 0.220 & \underline{0.293} & \underline{0.219} & 0.296 & \textbf{0.215} & \textbf{0.292} \\
 & 336 & 0.278 & \underline{0.330} & \underline{0.275} & 0.336 & \textbf{0.268} & \textbf{0.329} \\
 & 720 & 0.364 & \underline{0.385} & \underline{0.359} & 0.392 & \textbf{0.349} & \textbf{0.383} \\
\cmidrule(lr){2-8}
 & \textit{Avg} & 0.257 & \underline{0.316} & \underline{0.254} & 0.320 & \textbf{0.248} & \textbf{0.314} \\
\midrule
Electricity & 96 & \textbf{0.127} & \textbf{0.222} & 0.133 & 0.232 & \underline{0.130} & \underline{0.228} \\
 & 192 & \underline{0.146} & \textbf{0.241} & 0.149 & \underline{0.247} & \textbf{0.143} & \textbf{0.241} \\
 & 336 & 0.162 & \textbf{0.258} & \underline{0.161} & \underline{0.259} & \textbf{0.160} & 0.260 \\
 & 720 & 0.206 & \underline{0.296} & \underline{0.197} & 0.297 & \textbf{0.173} & \textbf{0.274} \\
\cmidrule(lr){2-8}
 & \textit{Avg} & \underline{0.160} & \underline{0.254} & \underline{0.160} & 0.259 & \textbf{0.152} & \textbf{0.251} \\
\midrule
Traffic & 96 & \textbf{0.355} & \textbf{0.252} & 0.378 & 0.273 & \underline{0.371} & \underline{0.267} \\
 & 192 & \textbf{0.380} & \textbf{0.259} & 0.391 & 0.277 & \underline{0.385} & \underline{0.274} \\
 & 336 & \textbf{0.386} & \textbf{0.262} & 0.402 & \underline{0.282} & \underline{0.398} & \underline{0.282} \\
 & 720 & \textbf{0.424} & \textbf{0.284} & 0.434 & 0.297 & \underline{0.425} & \underline{0.295} \\
\cmidrule(lr){2-8}
 & \textit{Avg} & \textbf{0.387} & \textbf{0.264} & 0.402 & 0.282 & \underline{0.395} & \underline{0.280} \\
\midrule
Solar & 96 & \textbf{0.162} & \textbf{0.218} & 0.192 & 0.251 & \underline{0.182} & \underline{0.244} \\
 & 192 & \textbf{0.177} & \textbf{0.236} & 0.247 & 0.323 & \underline{0.204} & \underline{0.254} \\
 & 336 & \textbf{0.185} & \textbf{0.241} & 0.240 & 0.300 & \underline{0.216} & \underline{0.263} \\
 & 720 & \textbf{0.196} & \textbf{0.250} & 0.246 & 0.311 & \underline{0.224} & \underline{0.270} \\
\cmidrule(lr){2-8}
 & \textit{Avg} & \textbf{0.180} & \textbf{0.236} & 0.231 & 0.296 & \underline{0.206} & \underline{0.258} \\
\bottomrule
\end{tabular}
\end{table}

\FloatBarrier

\section{Retrieval and Successor Evidence}
\label{app:placebo-retrieval}

We separate candidate selection from successor evidence through four
retrieval perturbations.
\emph{Random eligible} samples candidates from the valid search pool.
\emph{Bottom-$K$} selects the lowest scoring candidates. \emph{Nearest only}
removes diversity and keeps the highest scoring candidate. \emph{Successor
shuffle} preserves the selected candidates but permutes their observed
continuations. These perturbations isolate whether the correction benefits from
the learned ranking and from the trajectories that follow the retrieved
states.

\begin{table}[H]
\centering
\caption{Detail-retrieval perturbations that change candidate selection or
successor pairing. Bold marks the lowest MSE in each row.}
\label{tab:duots-placebo-retrieval}
\scriptsize
\setlength{\tabcolsep}{4pt}
\renewcommand{\arraystretch}{0.92}
\resizebox{\textwidth}{!}{%
\begin{tabular}{lc ccccc}
\toprule
Dataset & $H$ & Default & Random eligible & Bottom-$K$ &
Nearest only & Successor shuffle \\
\midrule
\multirow{4}{*}{ETTh1}
 & 96  & 0.3634 & 0.3666 & 0.3690 & 0.3634 & \textbf{0.3632} \\
 & 192 & 0.4009 & 0.4027 & 0.4080 & 0.4009 & \textbf{0.3989} \\
 & 336 & 0.4017 & 0.4017 & 0.4015 & 0.4017 & \textbf{0.4015} \\
 & 720 & \textbf{0.4412} & 0.4416 & 0.4416 & \textbf{0.4412} & 0.4413 \\
\midrule
\multirow{4}{*}{ETTh2}
 & 96  & \textbf{0.2632} & 0.2632 & 0.2632 & \textbf{0.2632} & 0.2634 \\
 & 192 & \textbf{0.3125} & 0.3125 & 0.3125 & \textbf{0.3125} & \textbf{0.3125} \\
 & 336 & 0.3144 & 0.3152 & \textbf{0.3087} & 0.3144 & 0.3130 \\
 & 720 & 0.3835 & 0.3829 & \textbf{0.3823} & 0.3834 & 0.3829 \\
\midrule
\multirow{4}{*}{ETTm1}
 & 96  & \textbf{0.2849} & 0.2851 & 0.2851 & \textbf{0.2849} & 0.2850 \\
 & 192 & \textbf{0.3340} & 0.3347 & 0.3346 & \textbf{0.3340} & 0.3350 \\
 & 336 & \textbf{0.3643} & 0.3654 & 0.3659 & \textbf{0.3643} & 0.3648 \\
 & 720 & \textbf{0.4193} & 0.4196 & 0.4202 & 0.4194 & 0.4213 \\
\midrule
\multirow{4}{*}{ETTm2}
 & 96  & 0.1665 & 0.1669 & 0.1677 & \textbf{0.1665} & 0.1675 \\
 & 192 & \textbf{0.2203} & 0.2206 & 0.2206 & \textbf{0.2203} & 0.2214 \\
 & 336 & \textbf{0.2789} & 0.2795 & 0.2798 & \textbf{0.2789} & 0.2797 \\
 & 720 & \textbf{0.3644} & \textbf{0.3644} & \textbf{0.3644} & \textbf{0.3644} & 0.3724 \\
\midrule
\multirow{4}{*}{Solar}
 & 96  & \textbf{0.1625} & 0.1704 & 0.1807 & \textbf{0.1625} & 0.1635 \\
 & 192 & \textbf{0.1778} & 0.1790 & 0.1800 & \textbf{0.1778} & 0.1781 \\
 & 336 & 0.1852 & 0.1853 & 0.1854 & 0.1852 & \textbf{0.1852} \\
 & 720 & \textbf{0.1968} & 0.1980 & 0.2010 & \textbf{0.1968} & 0.1969 \\
\midrule
\multirow{3}{*}{Weather}
 & 96  & 0.1428 & 0.1432 & 0.1444 & \textbf{0.1428} & 0.1429 \\
 & 192 & 0.1889 & 0.1905 & 0.1926 & 0.1889 & \textbf{0.1888} \\
 & 720 & 0.3126 & 0.3126 & \textbf{0.3126} & 0.3126 & 0.3126 \\
\bottomrule
\end{tabular}%
}
\end{table}

Random selection changes 26 settings and increases MSE in
25 of them, with a mean increase of 0.00081. Bottom-$K$ also changes 26
settings and worsens 22, with a mean increase of 0.00160. The consistent
degradation shows that the learned ranking provides useful retrieval evidence.
Successor shuffling changes 27 settings and increases MSE in 18, with a
smaller mean increase of 0.00040. This result indicates that the observed
continuations also contribute, although less uniformly. Nearest-only affects
only the five settings that use more than one candidate and shows no consistent
advantage. Overall, the correction depends mainly on meaningful candidate
selection and also benefits from the associated successor trajectories.

\section{Additional Mechanism Diagnostics}
\label{app:additional-analysis}

\paragraph{Learned correction scale.}
Across all 423 dataset--horizon--future-patch combinations, the mean absolute
correction scale is $0.040$. The horizon averages are $0.041$ on Solar,
$0.056$ on Weather, and $0.031$ on Traffic. The correction therefore remains
small relative to the normalized base forecast instead of replacing it. The
sign is not constrained. Eight of the 32 cells learn a negative scale at every
future patch.

\paragraph{Confidence-map calibration.}
The affine confidence maps do not move uniformly. Of 64 dataset--horizon--level
pairs, seven stay exactly at their zero initialization and 37 remain within
$0.05$ of it. In contrast, the parameters move substantially for the selected
Solar, Traffic, and Weather cases. For example, at $H=336$ the journey-level
$(s,b)$ parameters are $(-0.074,-0.629)$ on Solar and
$(-0.387,-0.603)$ on Traffic, while Weather learns nonzero maps at both levels.
This pattern indicates selective calibration rather than a uniformly active
confidence transform.

\paragraph{Where does the correction help?}
Table~\ref{tab:error-decomposition} compares the corrected and base forecasts
on Solar, Traffic, and Weather, where the correction improves average MSE at
every horizon. Weather improves at every forecast step. Solar's gain is
distributed across nearly all channels, and Traffic also improves across most
steps and channels. The decomposition therefore supports the correction's
benefit at both step and channel levels.

\begin{table}[H]
\centering
\caption{Correction decomposition averaged over four horizons. $\Delta$MSE is corrected minus base
($\times 10^{-3}$). The last columns report where the correction has larger
squared error.}
\label{tab:error-decomposition}
\small
\setlength{\tabcolsep}{7pt}
\renewcommand{\arraystretch}{1.05}
\begin{tabular}{lrrr}
\toprule
Dataset & $\Delta$MSE & Steps hurt (\%) & Channels hurt (\%) \\
\midrule
Solar   & -9.45  & 24.5 & 0.2 \\
Traffic & -1.25  & 12.9 & 12.4 \\
Weather & -17.50 & 0.0  & 16.7 \\
\bottomrule
\end{tabular}
\end{table}

\section{Computational Cost}
\label{app:computational-cost}

\begin{table}[H]
\centering
\caption{Computational cost with $L=512$ and $H=96$
on one RTX 4060 Laptop GPU. ETT uses batch size 32, Electricity uses batch
size 4, and Weather uses batch size 16. Smaller values are better.}
\label{tab:lc6-efficiency}
\scriptsize
\setlength{\tabcolsep}{3pt}
\renewcommand{\arraystretch}{1.0}
\begin{tabular*}{\textwidth}{@{\extracolsep{\fill}}llrrrrrr@{}}
\toprule
Dataset & Model & Params & FLOPs & FLOPs/$C$ & Epoch & Latency & Memory \\
 & & (M) & (G) & (G/ch.) & (s) & (ms/sample) & (MB) \\
\midrule
\multicolumn{8}{l}{\textit{ETTh2} ($C=7$, batch size 32)} \\
 & DLinear & 0.10 & 0.001 & 0.0001 & 0.59 & 0.003 & 21 \\
 & TimeKAN & 0.05 & 0.035 & 0.0050 & 8.54 & 0.410 & 554 \\
 & iTransformer & 0.95 & 0.021 & 0.0030 & 1.53 & 0.021 & 47 \\
 & PatchTST & 7.40 & 1.971 & 0.2816 & 10.96 & 0.348 & 573 \\
 & DuoTS (Ours) & 12.97 & 2.648 & 0.3783 & 34.95 & 0.942 & 2057 \\
\midrule
\multicolumn{8}{l}{\textit{Electricity} ($C=321$, batch size 4)} \\
 & DLinear & 0.10 & 0.063 & 0.0002 & 12.10 & 0.044 & 37 \\
 & TimeKAN & 0.06 & 2.170 & 0.0068 & 1395.76 & 32.348 & 4180 \\
 & iTransformer & 5.05 & 3.918 & 0.0122 & 81.12 & 1.368 & 274 \\
 & PatchTST & 7.40 & 90.365 & 0.2815 & 1323.39 & 24.970 & 2729 \\
 & DuoTS (Ours) & 3.04 & 13.229 & 0.0412 & 1151.09 & 10.051 & 2783 \\
\midrule
\multicolumn{8}{l}{\textit{Weather} ($C=21$, batch size 16)} \\
 & DLinear & 0.10 & 0.004 & 0.0002 & 5.38 & 0.007 & 23 \\
 & TimeKAN & 0.06 & 0.173 & 0.0082 & 200.28 & 2.348 & 1408 \\
 & iTransformer & 1.21 & 0.061 & 0.0029 & 19.04 & 0.044 & 55 \\
 & PatchTST & 9.46 & 17.418 & 0.8294 & 486.50 & 3.438 & 3600 \\
 & DuoTS (Ours) & 1.09 & 0.268 & 0.0127 & 51.94 & 0.188 & 304 \\
\bottomrule
\end{tabular*}
\end{table}

Table~\ref{tab:lc6-efficiency} uses a common look-back window of 512, horizon
96, and a single RTX 4060 Laptop GPU. On Electricity, DuoTS has 3.04M
parameters and uses 13.229G FLOPs, compared with PatchTST's 7.40M parameters
and 90.365G FLOPs. On Weather, DuoTS lies between iTransformer and PatchTST in
FLOPs, epoch time, and latency. ETTh2 is the unfavorable case. Its seven
channels do not amortize the fixed wide-representation and correction costs,
so DuoTS is less efficient than every listed baseline.

\end{document}